# Persistent AUV Operations Using a Robust Reactive Mission and Path Planning ($R^{II}MP^{II}$) Architecture

**S. MahmoudZadeh**
School of Information Technology, Deakin University, Geelong, Victoria 3220, Australia
S.MahmoudZadeh@deakin.edu.au

**Abstract** Providing a higher level of decision autonomy and accompanying prompt changes of an uncertain environment is a true challenge of AUVs autonomous operations. The proceeding approach introduces a robust reactive structure that accommodates an AUV's mission planning, task-time management in a top level and incorporates environmental changes by a synchronic motion planning in a lower level. The proposed architecture is developed in a hierarchical modular format and a bunch of evolutionary algorithms are employed by each module to investigate the efficiency and robustness of the structure in different mission scenarios while water current data, uncertain static-mobile/motile obstacles, and vehicles Kino-dynamic constraints are taken into account. The motion planner is facilitated with online re-planning capability to refine the vehicle's trajectory based on local variations of the environment. A small computational load is devoted for re-planning procedure since the upper layer mission planner renders an efficient overview of the operation area that AUV should fly thru. Numerical simulations are carried out to investigate robustness and performance of the architecture in different situations of a real-world underwater environment. Analysis of the simulation results claims the remarkable capability of the proposed model in accurate mission task-time-threat management while guarantying a secure deployment during the mission.

## 1 Introduction

AUVs have exposed their capabilities in cost effective underwater explorations up to thousands of meters, far beyond what humans are capable of reaching, and today are the first choice to navigate autonomously and undertake various missions. AUVs are largely employed for various purposes such as inspection and survey (Marthiniussen et al. 2004), scientific underwater explorations (Iwakami et al. 2002, Sibenac et al. 2002), sampling and monitoring coastal areas (An et al. 2004), measuring turbulence and subsea data collection (Dhanak and Holappa 1996, Millard et al. 1998), and military applications (Djapic and Nad 2010). However, autonomous adaptation of AUVs in performing different tasks in dynamic and continuously changing environment has not yet been completely fulfilled, where it is still necessary for the operators to remain in the loop of considering and making decisions (Johnson et al. 2007). Hence, an advanced degree of autonomy at the same level as human operator is an essential prerequisite to trade-off within tasks importance and the problem constraints while adapting the terrain changes during the mission, which depends on the design of an accurate task allocation and motion planning scheme. Generally, autonomy for AUVs can be considered in three primary levels as follows:

- **First level:** the vehicle is adjusted and stabilized by feedback control and can autonomously route its trajectory to assigned waypoints. At this level, the vehicle is capable of locating its position on the map and follows the defined path, without operator's interaction.

- **Second level:** the vehicle's navigation system has capability of indicating intermediate waypoints, which allows the human operator to define main targets without giving details. At this level, vehicle has also capability of autonomous departure, return and collision avoidance.

- **Third level:** the human operator specifies high-level mission scenarios and the vehicle holds extensive mission-related information. At this level, the system also holds the functionalities of both above mentioned levels.

The third level of autonomy is targeted by this research, which is addressed through the designed multi-layered deliberative decision-making framework. Increasing the levels of autonomy leads to minimizing the reliance on the human supervisor. Such self-controlled operation tightly depends on having a robust mission-motion planning and accurate task allocation mechanism. Due to this importance, autonomous motion planning and vehicles routing/task-allocation approach have been topic of interest for many research frameworks over the last two decades. Many algorithms and methods have been employed for unmanned vehicles path planning/ routing problem in recent years like Dijkstra's, D* and A* algorithms (Carsten et al. 2006, Likhachev et al. 2005). Dijkstra's calculate all applicable paths between two points, therefore takes long time to get to the appropriate answer. The A* also categorized as a best-first heuristic search method and acts more efficiently because of its heuristic searching capability. On the other hand, the D* algorithm operates based on a linear interpolation-based strategy that allows frequently update of heading directions, however, it is computationally expensive in high-dimensional problems. Fast Marching (FM) algorithm is another approach suggested for path planning of an underwater vehicle in a in a time-variant terrain, which is accurate but again computationally more complex than A* (Petres et al. 2005). Later on, this method has been upgraded to heuristically guided FM* on the same application (Petres et al. 2007) that uses accuracy of FM and time efficiency of the A*; however, this method is also inefficient for real-time application due to use of linear anisotropic cost in handling the computational complexity. The heuristic grid-search algorithms use discrete state transitions and have expensive time complexity especially in large scales that makes them inappropriate for real-time motion planning. The vehicle mission planning and routing-task scheduling is also vastly investigated suggesting various strategies such as graph matching algorithm (Kwok et al. 2002), Tabu search algorithm (Higgins 2001), partitioning method (Liu and Shell 2012), simulated annealing (Chiang and Russell 1996), and branch and cut algorithm (Lysgaard et al. 2004). Zhu and Yang (2010) applied an upgraded SOM-based approach for multi-robots dynamic task allocation. Later on, Karimanzira et al. (2014) developed waypoint tracking scheme along with a behavior based controller for an AUV guidance in large scale terrain. The deterministic methods produce better quality solutions, however these algorithms require considerable computational efforts and tend to fail when the problem size grows. Therefore, they are not appropriate approaches for real-time routing applications, specifically when the operating graph is topologically complex. In contrast, the meta-heuristic methods take less computation time and obtains optimal or near optimal solutions quickly. Generally, these algorithms have low sensitivity to graph complexity, so search time increases linearly with incrementing the number of points (MahmoudZadeh et al. 2017-a, Besada-Portas et al. 2010). An evolutionary route planning approach has been proposed by (MahmoudZadeh et al. 2015) for AUV task assignment and risk management joint problem in a large scale static terrain and this approach was extended to be implemented on semi dynamic network by (MahmoudZadeh et al. 2016-a) taking the use of BBO and PSO algorithms. The abovementioned approaches particularly focused on task and target assignment and scheduling problems without incorporating modality of deployment. In general, the routing/mission scheduling strategies give a top view the operation area for an AUV in which the vast space is splitted into smaller beneficent sections for AUV's operation; however, these approaches do not deal with quality of deployment in terms copping ocean variability. Meta-heuristics also applied successfully in the scope of path planning. Zamuda and Sosa (2014) applied a Differential Evolution (DE) based path planner for underwater glider to opportunistic sampling of dynamic mesoscale ocean structures. Fu et al. (2012) developed a Quantum Particle Swarm Optimization (QPSO) based offline path planner for unmanned aerial vehicle's operation in a static known terrain; however, such an offline planner is not sufficient to cover real-world variations and uncertainties. Later on, this method has been improved by (Zeng et al. 2014) to an online path planning applying shell space decomposition method for AUV operation in dynamic marine environment. Yazdani et al. (2016, 2017) used an inverse dynamic in the virtual domain (IDVD) trajectory generating method for on-board realization of an AUV. Undoubtedly there is always a significant requirement for more efficient optimization techniques in autonomous vehicle's motion planning problem.

### 1.1 Existing Challenges of AUV Motion Planning and Task Allocation

The challenges and difficulties associated with the motion-mission planning can be synthesized form two perspective; Firstly, the functionality of the algorithms being utilized, and secondly the competency of the techniques for real-time applications. Additional to addressed problems with selected strategies in both planning realm, many technical challenges still remained unaddressed. Although various path planning techniques have been suggested for autonomous vehicles, AUV-oriented applications still have several difficulties when operating across a large-scale geographical area. The computational complexity grow exponentially with enlargement of search space dimensions. Also, accurate estimation of a large uncertain operation field far-off the vehicles sensor coverage is not practical and reliable. Even though available ocean predictive approaches operate well in small scales and over short time periods, they produce insufficiency of accuracy to current prediction over long time periods in larger scales, especially in cases with lower information resolution (Hollinger et al. 2013). Moreover, a massive load of data from terrain updates should be computed repeatedly, which is unnecessary as only awareness of the environment in proximity of the vehicle is sufficient such that it can perform reaction to environmental sudden changes. On the other hand, the path planning strategies are not designed for handling task-mission scheduling problems as the nature of these two approach are completely different, therefore, using a mission planning strategy is inevitable for handling this aspect of an autonomous operation.

### 1.2 Research Motivation and Contribution

Assuming that the tasks for a specific mission are distributed in different areas of a waypoint cluttered graph-like terrain, an accurate task organizer-routing strategy guides the vehicle toward the destination waypoint in the graph while taking the best use of a restricted battery time by efficient ordering of the tasks. On the other hand, the path planning deals with vehicles guidance from one point to another taking the ocean uncertainty and variability in to account. Respectively, this research contributes a *Reactive Robust Mission and Path Planning* ($R^{II}MP^{II}$) architecture, in which a set of well-coordinated functional modules of *Task Assign Mission Planner* (**TAMP**) and a local *Online Path Planner* (**OPP**) processing in a closed loop system to concurrently ensure that mission goals are satisfied and the AUV has a safe deployment in a severe underwater environment.

The proceeding paper provides a framework to accommodate both high-level decision making and low-level action generator that is able to simultaneously organize the mission timing and maneuverability of the vehicle in a cluttered and uncertain environment. The proposed architecture offers a degree of flexibility for employing divers sorts of meta-heuristic algorithms and can perfectly synchronize them to have a unique performance. It is advantageous to join two disparate prospective of vehicle's autonomy in high level task organizing and low level self/environment awareness in motion planning in a cooperative and concordant manner in which adopting diverse algorithms by these modules do not detriment the real-time performance of the system. Such a modular structure provides a reusable and versatile framework that is easily upgradable and is applicable for a broad group of autonomous vehicles such as unmanned aerial, ground or surface vehicles. Parallel execution of the TAMP and OPP speed up the computation process. The operating field for operation of the OPP is splitted into smaller spaces between pairs of waypoints; thus, re-planning a new trajectory requires rendering and re-computing less information. This leads problem space reduction and detracting the computational burden, which is another reason for fast operation of the proposed approach. The total operation time of the framework is in the range of seconds that is a remarkable achievement for autonomous interactive operations.

On the other hand, the most critical factor for preserving the stability and consistency of the proposed $R^{II}MP^{II}$ scheme is maintaining comparably fast operation for each component of the system (TAMP/OPP) to prevent any of them from dropping behind the others. Any such a delay disrupts the concurrency of the entire system, and adding NP computational time into the equation would itself render a solution suboptimal. Meta–heuristics are the fastest approach introduced for solving NP-hard complexity of these problems and have been shown to produce solutions close to the optimum with high probability that are advantaged to be implemented on a parallel machine with multiple processors, which speeds up their computation process (Roberge et al. 2013, MahmoudZadeh et al. 2016-b). Although the solutions proposed by any meta–heuristic algorithm do not necessarily correspond to the optimal solution, it is more important to control the time, and thus we rely on the previously mentioned ability of meta-heuristic algorithms, including Biogeography-Based Optimization (BBO), Ant Colony Optimization (ACO), Genetic Algorithm (GA), Firefly Algorithm (FA), Particle Swarm Optimization (PSO), and Differential Evolution (DE) algorithms as employed by TAMP and OPP to find correct and near optimal solutions in competitive time.

### 1.3 Paper Organization

The rest of the paper is organized as follows. The mission planning and task organizing approach is discussed in Section 2. This section also defines AUV path planning problem and research assumptions. Section 3 introduces the architecture designed in this study and explains its functionalities. This section also provides a brief overview of the applied metaheuristic algorithms, and their application on AUV mission and motion planning approach. Section 4 discusses about mechanism of the $R^{II}MP^{II}$ architecture and evaluation the introduced approach in satisfying different aspects of an autonomous mission planning objectives. Section 5 is the conclusion of this research.

## 2 Problem Definition

### 2.1 AUV Modeling and Specifications

A general kinematic model used for path planning of an AUV like REMUS vehicle is as follows (Fossen 2002):

$$\begin{aligned}
\dot{X} &= u\cos(\psi)\cos(\theta) - v\sin(\psi) + w\cos(\psi)\sin(\theta) \\
\dot{Y} &= u\sin(\psi)\cos(\theta) + v\cos(\psi) + w\sin(\psi)\sin(\theta) \qquad (1) \\
\dot{Z} &= -u\sin(\theta) + w\cos(\theta)
\end{aligned}$$

where $X$, $Y$, and $Z$ are the coordinates of the AUV's center of gravity in the North-East-Down frame represented by {n}; $u,v,$ and $w$ are surge, sway, and heave velocity components relative to the water in the body frame {b}; $\psi$ and θ are the yaw and pitch angles in the {n}-frame. It is noteworthy to mention that in (1) the roll angle is assumed negligible and is passively controlled. The origin of the {n}-frame is set at the surface, so that z represents the vertical distance from the surface, i.e. AUV depth. The components of the REMUS vehicle velocity are obtained as follows:

$$\begin{aligned}
u &= |\upsilon|\cos(\psi)\cos(\theta) + |V_c|\cos(\psi_c)\cos(\theta_c) \\
v &= |\upsilon|\sin(\psi)\cos(\theta) + |V_c|\sin(\psi_c)\cos(\theta_c) \qquad (2) \\
w &= |\upsilon|\sin(\theta) + |V_c|\sin(\theta_c)
\end{aligned}$$

where, $\upsilon$ represents the constant resultant velocity of the vehicle with respect to the ground, and $V_c$ expresses the magnitude of current disturbance in horizontal and vertical directions shown by $\psi_c$ and $\theta_c$ (see Eq.(8)). Finally, the REMUS vehicle selected for this study has the following specifications (AUV Systems, website):

- It is a modular AUV with dimensions of 0.2*m* in diameter, 1.6*m* in length, and less than 45*kg* in weight;
- It can operate with maximum forward velocity of approximately 5 *knots* for maximum mission distances of approximately 55 *km* and can operate up to the depth of 100 *m*;
- It is equipped with a long range Acoustic Doppler Current Profiler (ADCP) operating in 75 *kHz* and is able to measure currents profiles up to 1km ahead of the vehicle;
- It is equipped with upward looking downward looking, and sidescan sonar sensors operating in range of 75-540 *kHz* to percept and measure obstacles' features such as obstacles' coordinate and velocity;
- It is additionally equipped with underwater modem communication, GPS navigation, a Seabird CTD, a Wetlabs ECO sensor, and Wi-Fi.

### 2.2 AUV TAMP Problem

For any application of the AUV a set of tasks is predefined and initialized in advance with specific characteristics, then get dispatched to the vehicle in the format of commands. Totally 30 different tasks specified in this research, where each task is assigned with a specific weight initialized once in advance using a normal distribution given by Eq.(3). A three-dimensional terrain ($\Gamma_{3D}$) based on real map of the Whitsunday Islands is modelled for the purpose of this research to resemble realistic marine environment. Tasks for a specific mission are distributed in eligible water covered sections of the map, in which placement of the tasks is mapped and presented in a graph format where beginning and ending location of a task appointed with waypoints. Accordingly, the operating area is mapped with undirected connected weighted network denoted by $G = (P, E)$, where $P$ is the vertices of the graph that corresponds to waypoints and $E$ denotes the edges of the graph in which some of the edges assigned with a specific task that formulated as follows:

$$\aleph = \{\aleph_1, \ldots, \aleph_{30}\}; \quad \forall \aleph_i, \quad w_i \sim \mathbf{N}(20,10) \tag{3}$$

$$G = (P, E) \Rightarrow \begin{matrix} |P| = k \\ |E| = m \end{matrix} \Rightarrow \begin{matrix} P(G) = \{p^1, \ldots, p^k\} \\ E(G) = \{e^1, \ldots, e^m\} \end{matrix} \Rightarrow e^{ij} = \left(p^i, p^j\right)$$

$$\Gamma_{3D} : 10000_x \times 10000_y \times 100_z \tag{4}$$

$$\forall p^i \in P \;\; \Rightarrow \begin{matrix} p^i_{x,y} \sim \mathbf{U}(0,10000) \\ p^i_z \sim \mathbf{U}(0,100) \end{matrix} \Rightarrow p^i_{x,y,z} \notin \{MAP = 0\}$$

$$\forall e^{ij} \;\; \exists \;\; w_{ij}, d_{ij}, t_{ij} \Rightarrow \begin{cases} w_{ij} > 1 & if \;\; \exists \;\; \aleph_{ij} \\ w_{ij} = 1 & if \;\; \exists! \;\; \aleph_{ij} \end{cases}$$

$$d_{ij} = \sqrt{\left(p^j_x - p^i_x\right)^2 + \left(p^j_y - p^i_y\right)^2 + \left(p^j_z - p^i_z\right)^2} \tag{5}$$

$$t_{ij} = {d_{ij}} \big/ {|\upsilon|}$$

where $\aleph$ denotes the tasks and $w$ is the task importance, all waypoints are connected but only the edges ($e^{ij}$) that assigned with a task are weighted with a value more than one that is calculated based on attributes of the corresponding task. The $d_{ij}$ and $t_{ij}$ are distance along the $e^{ij}$ and time for travelling the $e^{ij}$. In the given relations, $\{Map=0\}$ corresponds to the coastal area of map which is defined as the forbidden area for vehicles deployment and $|\upsilon|$ is the AUV water referenced velocity. The water covered eligible sections of the map is identified using an efficient k-means clustering method, in which the blue sections of the map is recognized as the water and the rest are categorized as coastal or uncertain zones according to its color range. Example of such a graph like terrain is given by Fig.1.

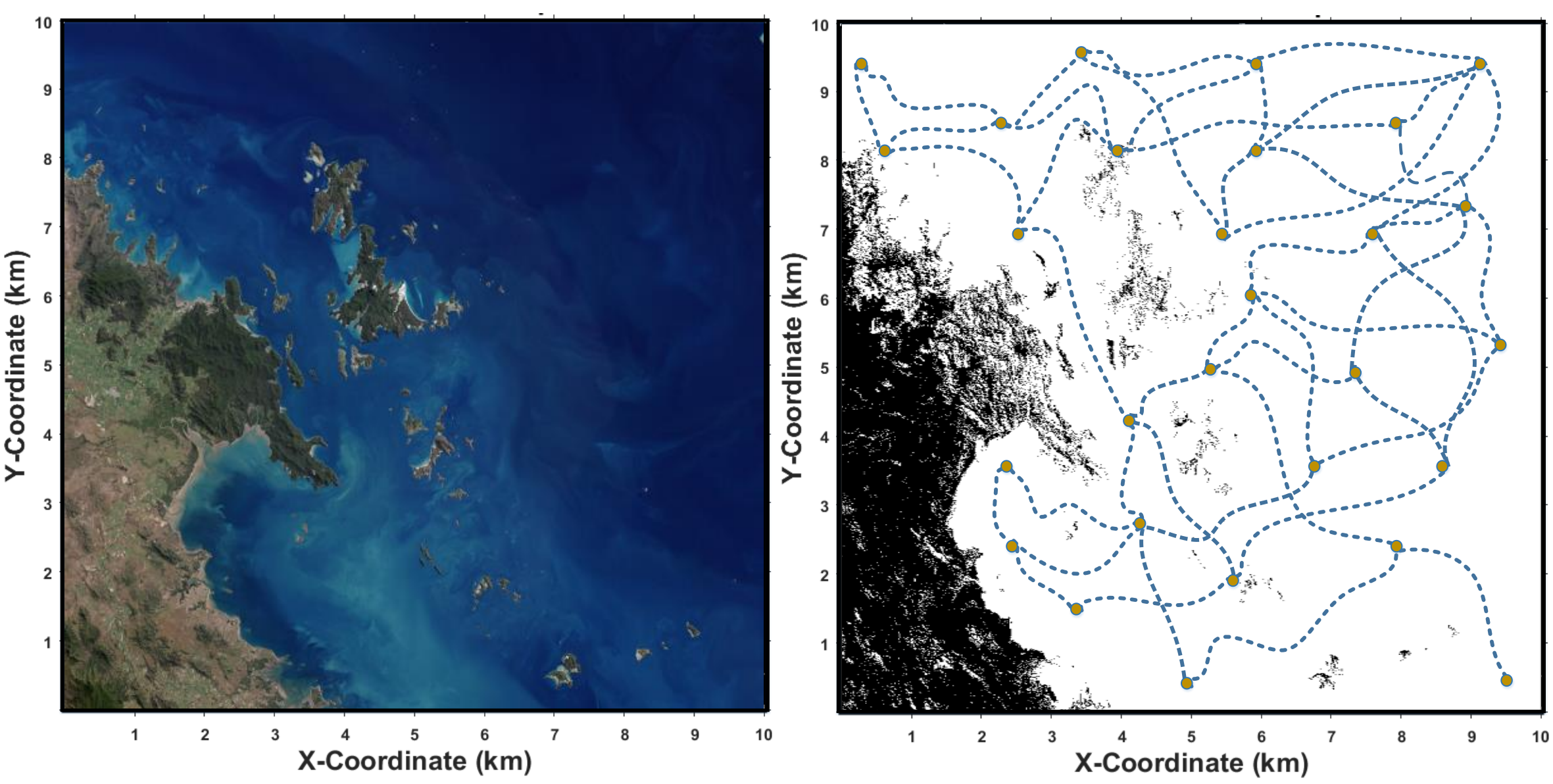

Fig.1. The original and clustered map of the Whitsunday Islands and graph representation of operating area

In such a terrain that covered by several waypoints/tasks, the vehicle is requested to complete maximum possible number of tasks with highest weight in the total available time denoted by $T_\tau$. Hence, the TAMP tends to find the best fitted route to $T_\tau$ collecting best sequence of vertices in the graph in a way that total weigh of route is maximized which means the edges with best tasks are selected in a manner to guide the vehicle to destination vertex while on-time termination of the journey is assured similar to TSP-Knapsack problems. In the preceding discussion, this problem is described as follows:

$$\Re_k = \left(p^S_{x,y,z}, \ldots, p^i_{x,y,z}, \ldots, p^D_{x,y,z}\right)$$

$$\forall e^{ij} = \left(p^i_{x,y,z}, p^j_{x,y,z}\right) \;\; \exists \;\; w_{ij}, d_{ij}, t_{ij} \Rightarrow \begin{cases} w_{ij} > 1 & if \;\; \exists \;\; \aleph_{ij} \\ w_{ij} = 1 & f \;\; \exists! \;\; \aleph_{ij} \end{cases} \tag{6}$$

$$
\begin{aligned}
& T_{\Re} = \sum_{\substack{i=0 \\ j\neq i}}^{n} s_{eij} \times t_{ij} = \sum_{\substack{i=0 \\ j\neq i}}^{n} s_{eij} \times \left( {d_{ij}} \middle/ {|\upsilon|} \right), \quad s_{eij} \in \{0,1\} \\
& C_{\Re} \propto \left| T_{\Re} - T_{\tau} \right| + \left( 1 \middle/ \sum_{\substack{i=0 \\ j\neq i}}^{n} s_{eij} \times w_{ij} \right) \\
& s.t. \\
& \forall \Re_i \Rightarrow \max(T_{\Re}) < T_{\tau}
\end{aligned} \tag{7}
$$

where, the $\Re$ is an arbitrary route commenced with $p^s$ and ended at $p^D$, $s_{eij}$ is the selection that equals to 1 for the selected and 0 for unselected edges. $T_{\Re}$ is the route time that should approach but not overstep the $T_{\tau}$. Hence, the problem is a restricted multi-objective optimization problem and $C_{\Re}$ denotes the route cost. The generated route should be feasible to following criteria:

1) Any feasible $\Re$ is commenced and ended with the index of the start and destination vertex in the graph.
2) The feasible $\Re$ excludes the non-existent edges in the graph.
3) The feasible $\Re$ doesn't meet a specific node for multiple times.
4) The feasible $\Re$ doesn't pass an edge more than once.

To fulfil the objectives of the proposed TAMP, four popular meta-heuristic optimization algorithms of the GA, PSO, ACO, and BBO are applied that perform promising performance in solving NP hard problems. Accurate coding and assigning the individual population is the most important step in implementing all evolutionary algorithms; thus, in this research, the individual populations for all algorithms are initialized with feasible and valid routes $\Re$, in which adjacency matrix information a random priority vector is employed for feasible route generation (MahmoudZadeh et al. 2016-c, 2016-d).

### 2.3 AUV Path Planning Problem

A necessary requirement for vehicles safe and accurate deployment is designing a proper path planning scheme to cover objectives of the current research in lower level of autonomy. The path planning aims to guide the vehicle between two specific points with minimum time due to battery limitation while ensuring safe deployment coping various real world disturbances, which is known as an NP-hard optimization problem. The path planner designed in the proceeding research is capable of extracting eligible areas of map for vehicles deployment, carrying out obstacle avoidance and coping water current disturbance. Current can have desirable or disturbing effect on vehicles deployment as the desirable current can drive the vehicle toward the destination which cause a remarkable save in battery usage, so the current is a critical factor that can affect optimality of the path. The current map in this research is captured from superposition of multiple viscous Lamb vortices (Garau et al. 2006) and two dimensional Navier-Stokes equation that forecast ocean conditions in general; hence, the physical model of current dynamics employed by the AUV is described by:

$$
\begin{aligned}
& u_c(\vec{S}) = -\Im \frac{y - y_0}{2\pi(\vec{S} - \vec{S}^O)^2} \left[ 1 - e^{\frac{-(\vec{S}-\vec{S}^O)^2}{\ell^2}} \right] \\
& v_c(\vec{S}) = \Im \frac{x - x_0}{2\pi(\vec{S} - \vec{S}^O)^2} \left[ 1 - e^{\frac{-(\vec{S}-\vec{S}^O)^2}{\ell^2}} \right] \\
& \vec{V}_c = (u_c, v_c) : \begin{cases} u_c = |V_c| \cos\theta_c \cos\psi_c \\ v_c = |V_c| \cos\theta_c \sin\psi_c \end{cases}
\end{aligned} \tag{8}
$$

The $S=(x,y)$ is a 2D space, $S^o$ is the center of the vortex, and $\ell$ is the radius of the vortex, and $\Im$ is the strength of the vortex, $V_C$ is magnitude of current speed, $\psi_c$ and $\theta_c$ are directions of current in horizontal and vertical planes, respectively. In this research, the vehicle's position $X, Y, Z$ along the generated potential path $\wp_i$ is generated using B-Spline curves captured from a set of control points $\vartheta$:$\{\vartheta^1_{x,y,z},\ldots,\vartheta^i_{x,y,z},\ldots,\vartheta^n_{x,y,z}\}$. Control points should be located in eligible areas of joint water sections in the clustered map within the respective search region between start and target points constraint to predefined upper and lower bounds of $\vartheta \in [U^i_{\vartheta}, L^i_{\vartheta}]$ in Cartesian coordinates. Therefore, the path $\wp_i$ and vehicles angular velocities are defined using following relation:

$$
\begin{aligned}
& \psi = \tan^{-1}\left( \frac{\left| \vartheta_y^{i+1} - \vartheta_y^{i} \right|}{\left| \vartheta_x^{i+1} - \vartheta_x^{i} \right|} \right) \\
& \theta = \tan^{-1}\left( \frac{-\left| \vartheta_z^{i+1} - \vartheta_z^{i} \right|}{\sqrt{\left( \vartheta_y^{i+1} - \vartheta_y^{i} \right)^2 + \left( \vartheta_x^{i+1} - \vartheta_x^{i} \right)^2}} \right) \\
& \langle X(t), Y(t), Z(t) \rangle \approx \left\langle \sum_{i=1}^{n} \left( \vartheta_x^i(t), \vartheta_y^i(t), \vartheta_z^i(t) \right) \right\rangle \\
& \wp^i_{x,y,z} = \sum_{i=1}^{|\wp|} \sqrt{(\vartheta_x^{i+1} - \vartheta_x^i)^2 + (\vartheta_y^{i+1} - \vartheta_y^i)^2 + (\vartheta_z^{i+1} - \vartheta_z^i)^2} \\
& \wp(t) = [X(t), Y(t), Z(t), \psi(t), \theta(t), u(t), v(t), w(t)]
\end{aligned} \tag{9}
$$

The water referenced velocity of the vehicle is assumed to be constant, so the battery usage is a constant function of the time and distance travelled. Another factor for validation of the path planner is its capability in satisfying environmental and vehicular constraints. The environmental constraints are associated with the vehicles depth limitation, forbidden zones of map or intersecting any obstacle, and coping current flow that may causes drift between desired and actual deployment of the vehicle. The vehicular constraints also defined in a specific ranges for vehicle's surge, sway and yaw parameters. Obstacles are modelled to be static or moving based on previous research (MahmoudZadeh et al. 2016-c). The path cost is calculated by

$$
\begin{aligned}
&\forall \wp_{x,y,z} \Rightarrow \quad T_{\wp} = \sum_{1}^{|\wp|} \Gamma_{3D}\left\{ \vartheta_{t_i}^{i} \right\} = \sum_{1}^{|\wp|} \frac{\left| \vartheta_{i+1}^{t_{i+1}} - \vartheta_{i}^{t_i} \right|}{|\upsilon|} \\
&\nabla_{\Sigma_{M,\Theta}} = \begin{cases} 1 & \wp_{x.y.z}(t) = Coast : Map(x,y) = 1 \\ 1 & \wp_{x.y.z}(t) \cap \bigcup_{N\Theta} \Theta(\Theta_p, \Theta_r, \Theta_{Ur}) \\ 0 & Otherwise \end{cases} \\
&\nabla_{\wp} = \varepsilon_{z\min} \times \min\left(0; Z(t) - Z_{\min}\right) + \varepsilon_{z\max} \times \max\left(0; Z(t) - Z_{\max}\right) + \varepsilon_u \times \max\left(0; u(t) - u_{\max}\right) + \ldots \qquad (10) \\
&\quad \ldots + \varepsilon_v \times \max\left(0; |v(t)| - v_{\max}\right) + \varepsilon_{\psi} \times \max\left(0; |\dot{\psi}(t)| - \dot{\psi}_{\max}\right) + \varepsilon_{\Sigma_{M,\Theta}} \times \nabla_{\Sigma_{M,\Theta}} \\
&C_{\wp} = T_{\wp} + \sum_{i=1}^{n} Q_i f\left(\nabla_{\wp}\right)
\end{aligned}
$$

$Q_i\, f(\nabla_{\wp})$ is a weighted violation function that respects the AUV kinodynamic and collision constraints including depth violation ($Z$), to prevent the path from straying outside the vertical operating borders, surge ($u$), sway ($v$), yaw ($\psi$) violations, and the collision violation ($\nabla_{\Sigma M,\Theta}$) specified to prevent the path from collision danger. The $\varepsilon_{zmin}$, $\varepsilon_{zmax}$, $\varepsilon_{u}$, $\varepsilon_{v}$, $\varepsilon_{\psi}$, and $\varepsilon_{\Sigma M,\Theta}$ respectively denote the impact of each constraint violation in calculation of total path violation, and $C_{\wp}$ is the path cost. Considering the given relations by Eq.(9)-(10), appropriate adjustment of the control points play a substantial role in optimality of the generated trajectory. To this purpose, four popular meta-heuristic algorithms of PSO, BBO, DE, and FA are conducted by the proposed OPP that perform promising performance in solving NP hard problems and particularly motion planning problems.

## 3 R$^{II}$MP$^{II}$ Architecture

Figure 2 illustrates the R$^{II}$MP$^{II}$ architecture developed in the proceeding research. This modular framework is an upgraded version of the recently introduced model (MahmoudZadeh et al. 2017-b), where the modules functionality is improved by realistic modelling of the environment and adding re-planning capability to TAMP and OPP. The main goal of the R$^{II}$MP$^{II}$ model is to select the best sequence of tasks in a restricted time so that the vehicle is guided toward the final destination while guaranteeing on time termination of the mission and safe deployment during the mission. Assuming that the tasks for a specific mission are distributed in different areas of a waypoint cluttered graph-like terrain, the TAMP facilitates the vehicle to maximize the weight of the selected edges, which means selecting the best sequence of highest priority tasks in a limited time. The TAMP splits the operating area for OPP to distance between pairs of waypoints. Then OPP deals with vehicle's guidance from one point to another taking the ocean uncertainty and variability in to account. The local path cost of $C_{\wp}$ has a proportional relation to the travelled distance/time between pairs of vertices in the graph. The OPP operates concurrently in background of the TAMP; hence, the cost of generated local path $C_{\wp}$ has direct impact on route cost of $C_{\mathfrak{R}}$ and consequently the mission cost. Thus, the mission cost $C_{\mathrm{M}}$ is defined as a combination of multiple weighted cost functions that should maximized or minimized, in which the model is searching for a beneficent solution in the sense of the best combination of task, path, and route cost as given by Eq.(11). During the path planning process, traversing the corresponding distance may consume more time than what expected. Restitution of the wasted time necessitates a mission re-planning process, in which the re-routing is performed to compute the new collection of tasks fitted to the updated available time. The mission re-planning criteria is investigated after any completion of the OPP process (when a waypoint in the route sequence is visited). Hence, a computation time also is encountered in calculation of the total mission cost as given below:

$$
\begin{aligned}
&d_{ij} \propto \wp \quad \& \quad t_{ij} \propto T_{\wp} \\
&T_{\mathfrak{R}} = \sum_{\substack{i=0 \\ j \neq i}}^{n} s_{eij} \times t_{ij} \\
&C_{\mathfrak{R}} = \left| \sum_{\substack{i=0 \\ j \neq i}}^{n} s_{eij} \times \left( C_{\wp ij} \right) - T_{\tau} \right| \qquad (11) \\
&C_{\mathrm{M}} = \Phi_1 \left| \sum_{\substack{i=0 \\ j \neq i}}^{n} s_{eij} \times \left( C_{\wp ij} \right) - T_{\tau} \right| + \Phi_2 \left( 1 \Big/ \sum_{\substack{i=0 \\ j \neq i}}^{n} s_{eij} \times w_{ij} \right) + \sum_{1}^{r} T_{compute}
\end{aligned}
$$

where $\Phi_1$ and $\Phi_2$ are coefficients determine participation of each factor in calculation of the mission cost $C_{\mathrm{M}}$, $T_{compute}$ is the mission re-planning computation time, and $r$ is the number of re-planning procedure in a mission.

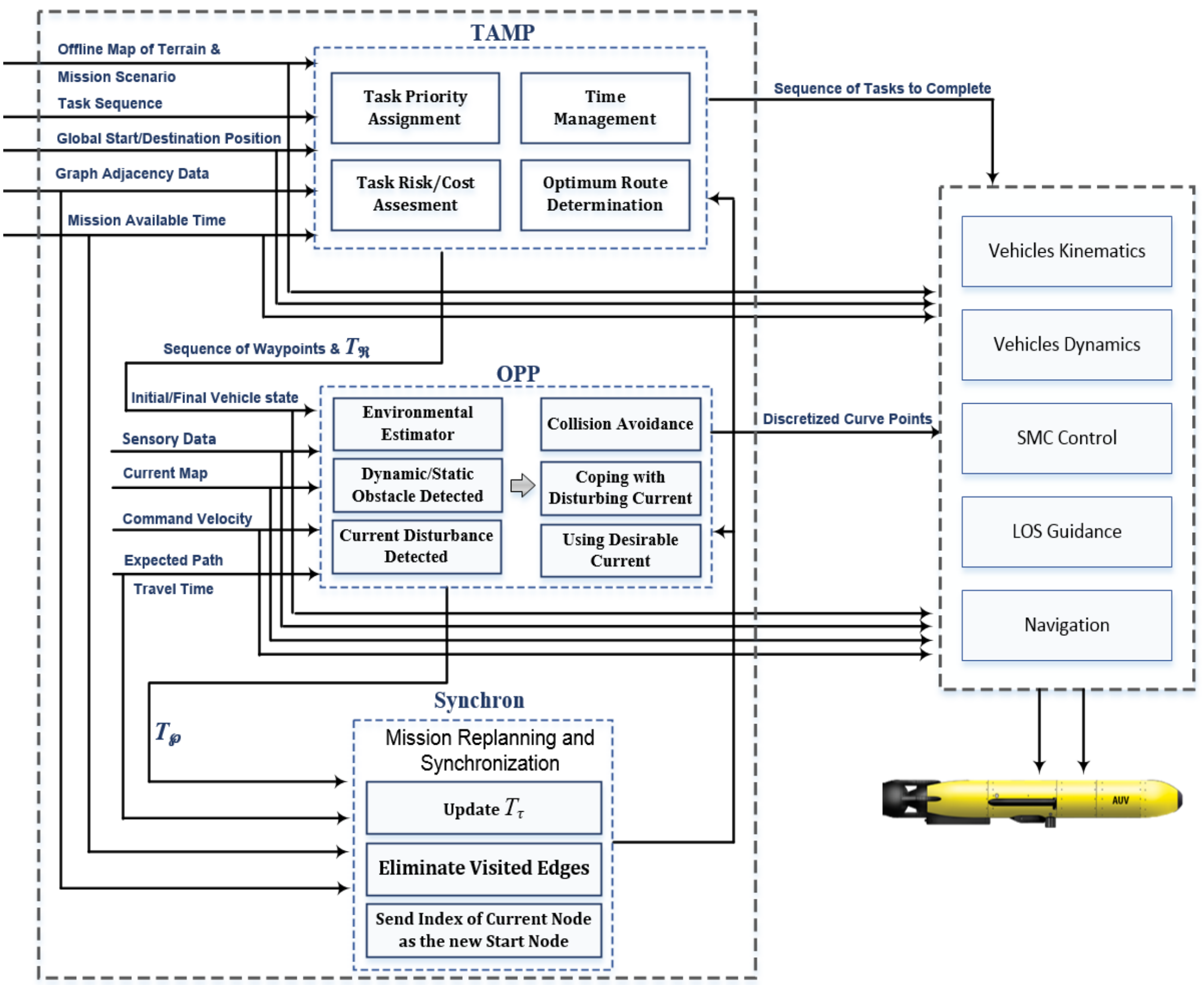

**Fig.2.** The operation diagram of the proposed R$^{II}$MP$^{II}$ model for autonomous operation of an AUV

The OPP module in this architecture is implemented using four evolutionary algorithms of BBO, PSO, DE, and FA, and evaluated and discussed in Section 4. The RMP approach is also evaluated through the Monte Carlo simulation for BBO, PSO, ACO and GA algorithms and discussed in detail by Section 4. All applied metaheuristic algorithms for the purpose of OPP and RMP have two inner loops through the population size of $i_{max}$ and iteration $t_{max}$ (refer to pseudo-codes); thus, in the worst case the computational complexity of the algorithms is O($i_{max}^2 \times t_{max}$) and the computation cost is respectively low as its complexity is linear to time. The cost evaluation is the most computationally complex part of almost all optimization problems. For evaluating the entire R$^{II}$MP$^{II}$ model a various combination of aforementioned algorithms are investigated and statistically analyzed through the series of Monte Carlo simulation.

## 3.1 Applied Meta-Heuristics in R$^{II}$MP$^{II}$ Architecture

### 3.1.1 Overview of the Employed Algorithms by TAMP Scheme

In fact, there is a significant distinction between theoretical understanding of metaheuristics and peculiarity of different applications in contrast. Specifically, this gap is more highlighted when scale (size), complexity, and nature of the problem is taken to account. Application of different methods may result very diverse on a same problem due to specific nature each problem. Therefore, accurate selection of the algorithm and proper matching of the algorithms functionalities and characteristics in accordance with nature of a particular problem is highly critical issue that should be taken into account in advance. These facts usually remained unattended and unsolved in most of the engineering and robotics frameworks in both theory and practice. Oftentimes metaheuristic algorithms require careful and accurate modifications to appropriately accommodate a specific problem. With respect to the combinatorial nature routing and mission planning problem, which is analogous to both Knapsack and Traveler Salesman Problems (TSP), this problem is categorized as a multi-objective NP-Hard problem often solved by optimization algorithms (MahmoudZadeh et al. 2016-b). Obtaining the optimal solutions for NP-hard problems is computationally challenging issue and difficult to solve in practice. Moreover, obtaining the exact optimum solution is only possible for the certain cases that the environment is completely known and no uncertainty exist; however, the modelled environment by this research corresponds to a spatiotemporal time varying environment with high uncertainty. Meta-heuristics are appropriate approaches suggested for handling above mentioned complexities.

The ACO algorithm has been introduced by Dorigo & Di Caro (1999) inspiring the ants' foraging process, and implemented successfully on TSP problem for the first time. This algorithm is proven to have a strong capability on solving vehicle routing problems and is well scaled with task assignment approaches due to its discrete nature (Dong et al. 2014, Euchi et al. 2015, Pendharkar 2015). On the other hand, the PSO is one of the fastest optimization methods for solving variety of the complex problems widely used in several studies in past decades. The particular problem of PSO is that it is operating in a continuous space originally which is in contrast with discrete nature of the search space in TAMP; however, the argument for using PSO strong enough as it scales well with complexity of multi-objective problems. This issue resolved by using the priority based route generation approach explained earlier (MahmoudZadeh et al. 2016-b), where this issue also increase the speed of the algorithm in finding optimum solution and prevent stucking in a local optima. GA is also a stochastic search algorithm operates based on biological evolution and has been extensively studied and widely used on different realms of graph routing problems and similar applications (Sharma et al. 2012). The GA also has a discrete nature and is well fitted with proposed problem. The BBO is another evolutionary technique developed according to equilibrium theory of island biogeography concept (Simon 2008). The population of candidate solutions in BBO is defined by geographically isolated islands known as habitat. A special feature of the BBO algorithm is that the original population never get discarded but get modified by migration, which this issue promotes the exploitation ability of the algorithm.

*Application of ACO on Mission Planning Approach*

In ACO, the ants find their path through the probabilistic decision according to level of pheromone concentration, so that the pheromone trail attract ants and guide them to grain source. Respectively, more pheromone is released on the overcrowded paths, which is a self-reinforcing process in finding the quickest route to target. The pheromone gets vaporized over time to helps ants to eliminate bad solutions that promotes the algorithm to keep the solutions from falling to the local optima. The pheromone concentration on edges (τ) keep the ant's collected information within the search process. The probability of targeting a node to travel is calculated by Eq.(12) and then, the total pheromone trail gets updated for all solutions according to Eq.(13).

$$p_{ij}^{k} = \begin{cases} \dfrac{\left(\tau_{ij}^{(k)}\right)^{\alpha}\left(\eta_{ij}^{(k)}\right)^{\beta}}{\sum\limits_{l \in N_l^k}\left(\tau_{il}^{(k)}\right)^{\alpha}\left(\eta_{il}^{(k)}\right)^{\beta}} & j \in N_i^k \\ 0 & j \notin N_i^k \end{cases} \tag{12}$$

$$\begin{aligned} &\tau_{ij} = (1-r)r + \sum_{k=1}^{N} \Delta\tau_{ij}^{(k)} \\ &\tau_{ij}(t+1) = \tau_{ij}(t) + \Delta\tau_{ij} \\ &\Delta\tau_{ij} = \begin{cases} -\lambda\tau_{ij}(t) + \dfrac{Q}{\Omega} & \textit{For nodes of the best ant} \\ -\lambda\tau_{ij}(t) & \textit{Otherwise} \end{cases} \end{aligned} \tag{13}$$

here, $k$ is the ant index, $i$ and $j$ denote the index of current and the target nodes, $\tau^{k}_{ij}$ is the pheromone concentration on edge between nodes $i$ and $j$, α and β are the pheromone and heuristic factors, respectively. $\eta_i$ denotes the heuristic information in $i^{th}$ node. $N_i^{(k)}$ indicates the set of neighbor of ant $k$ when located at node $i$. The $r \in (0,1]$ is evaporation rate. The $\Delta\tau_{ij}^{(k)}$ denotes the pheromone level collected by the best ant $k$ based on priority of ants released pheromone (Ω) on the selected nodes. $\tau^{k}_{ij}(t)$ and $\tau^{k}_{ij}(t+1)$ are the previous and current level of the pheromone between node $i$ and $j$. $Q$ is the amount of released pheromone on nodes taken by the best ant(s), and λ is the coefficient of pheromone evaporation. The procedure of the ACO based route planner is described by Fig.3.

*Application of GA on Mission Planning Approach*

The chromosome population in GA is initialized with feasible routes (task sequences); hence, the first and last gene of the chromosomes always corresponds to index of the start and destination nodes. New generation is produced from the initial population applying the GA operators of selection, crossover and mutation. Afterward, new generation gets evaluated by defined route feasibility criterion and the route cost function. The best fitted chromosomes with minimum cost get transferred to the next generation and the rest get eliminated. This process repeats iteratively until the population get converged to the best fitted solution to the given problem (Sivanandam and Deepa 2008). The average fitness of the population gets improved at each iteration by adaptive heuristic search nature of the GA. The operation is terminated when a fixed number of iterations get completed, or when no dramatic change observed in population evolution.

**[1]** ***Selection*:** Selecting the parents for crossover and mutation operations is another important step of the GA algorithm that improves the average fitness of the population in the next generation. This research uses the roulette wheel selection, wherein the next generation is selected from the best fitted chromosomes and the wheel divided into a number of slices so that the chromosomes with the best cost take larger slice of the wheel.

**[2]** ***Crossover*:** This operator shuffles sub parts of two parent chromosomes and generate mixed offspring. Generally, two main categories of single and multi-point crossover methods are introduced. Discussion over which crossover method is more appropriate still is an open area for research. This research applies uniform crossover, which uses fixed mixing ratio among parent chromosomes that usually is set on 0.5. The uniform crossover is remarkably suitable for large space problems. After offspring are generated, the new generation should be validated.

**[3]** ***Mutation*:** This operator flips multiple gents of a chromosome to generate new offspring. Current research used three types of insertion, inversion and swapping mutation techniques, in which all of them preserve most adjacency information. The mutation applied on gens between the first and last gens of the parent chromosome to keep the new generation in feasible space and then new offspring get evaluated.

The GA process gets terminated if maximum number of iteration is completed, or if population fitness doesn't change after several iterations and approach to a stall generation. The feasibility of the generated route is checked during the process of cost evaluation. The process of the GA algorithm on TAMP approach is proposed by Fig.3.

**ACO in TAMP Framework**

Initialization phase:

- Initialize population of ants in a colony using generated feasible routes
- Define number of layers in network, where each layer contains $p$ nodes i.e. $(x_{i1}, x_{i2}, \ldots, x_{ip})$
- Set initial pheromone for each edge by $\tau_{ij}^{(l)}=1$
- Set the maximum number of iterations $t_{max}$

***For*** $t=1$ ***to*** $t_{max}$
    ***For*** $k=1$ ***to*** $i_{max}$
        Compute the probability $P_{ij}$ of selecting node $j$ from the current node $i$ by Eq.(12).

- Assume $\alpha=1$
- Generate $k$ random numbers $r_k \in [0,1]$, one for each ant.
- Calculate range of the cumulative probability associated each routes applying roulette-wheel selection.
- Determine the route captured by ant $k$ as a cumulative probability that keeps $r_k$ random number in it.
- Evaluate the cost value for all solutions $f_k=f(x_t^{(k)})$; $k=1,2,\ldots,n$ and determine best and worst solutions $x_t^{best}$, $x_t^{worst}$
- Check the route (solution) feasibility criterion
- Save the best solution $x_t^{best}$

        Update the pheromone on arc when all the ants complete their tour using Eq.(13).
        ***For*** $j=1$ ***to*** $k$

$$\tau_{ij}^{k} = \tau_{ij}^{old} + \sum_{k=1}^{i_{\max}} \Delta\tau_{ij}^{k}$$

$$\tau_{ij}^{old} = (t-\rho)\tau_{ij}^{t-1}$$

            ***If*** $f(x_t^{best})^{new} < f(x_t^{best})^{old}$
                $x^{Globalbest}=(x_t^{best})^{new}$
            **else**
                $x^{Globalbest}=(x_t^{best})^{old}$
            ***end (if)***
        ***end (For)***
    ***end (For)***
    $\alpha_t=0.99\times\alpha_{t-1}$
    $\beta_t=0.99\times\beta_{t-1}$
***end (For)***
Output result

**GA in TAMP Framework**

Initialize the chromosome population by generated feasible routes

- Choose appropriate parameters for the population size $i_{max}$
- Set the maximum number of generations (iteration $t_{max}$)
- Assign maximum crossover and mutation coefficients.

***For*** $k=1$ ***to*** $t_{max}$
    Evaluate feasibility of each chromosome
    Calculate the fitness of each chromosome by defined cost function
    ***For*** $i=1$ ***to*** $i_{max}$
        Select best two chromosomes as parents using roulette wheel
        Apply the crossover operator
        Apply the mutation operator
        Check feasibility of off springs
        Evaluate fitness of offspring and select best between parents and offspring
    ***end (For)***
    Select the best fitted chromosomes to transfer to next generation
    Eliminate the worst chromosomes with highest cost value
***end (For)***
Output the most fitted choromosome as the best route (waypoint sequence)

**BBO in TAMP Framework**

Initialize a set of solutions as initial habitat population by generated feasible routes

- Choose appropriate parameters for the population size $i_{max}$
- Set the maximum number of generations (iteration $t_{max}$)
- Assign maximum immigration and emigration rate $(I, E)$
- Assign maximum mutation rate $m(S)$
- Set $S_{max}$ and SIV vector

***For*** $k=1$ ***to*** $t_{max}$
    Compute immigration rates λ and emigration rate μ for each solution
    Evaluate the fitness (HSI) of each habitat and identify Elite Habitats based on HIS
    Modify habitats based on λ and μ (Migration):
    ***For*** $i=1$ ***to*** $i_{max}$
        Use $\lambda_i$ to probabilistically decide whether immigrate to habitat $h_i$
        ***if*** $rand(0,1) < \lambda_i$
        ***For*** $j=1$ ***to*** *nPop*
            Select the emigrating habitat $h_j$ with probability $\propto \mu_j$
            **if** $rand(0,1) < \mu_j$
                Replace a randomly selected SIV variable of $h_i$ with its corresponding value in $h_j$
            ***end (if)***
        ***end (For)***
        ***end (if)***
    ***end (For)***
    Carry out the mutation based on probability
    Check the route Feasibility Criterion
    Implement the elitism to retain the best solution in the population from one generation to the next
***end (For)***
Output the best habitat with optimal fitness value and its corresponding route

**PSO in TAMP Framework**

Initialize swarm population with random velocity and position in following steps:

- Assign particle position $\chi_i$ with priority vector defined for route vector.
- Initialize each particle with random velocity $\upsilon_i$ in range of [-100,100]
- Choose appropriate parameters for the population size $i_{max}$
- Set the maximum number of iterations $t_{max}$
- Set the $\chi_i^{P\text{-}best}(1)$ and $\chi^{G\text{-}best}(1)$ with particle current position and swarm best at iteration $t=1$

***For*** $t=1$ ***to*** $t_{max}$
    Evaluate each candidate particle according to given cost function
    ***For*** $i=1$ ***to*** $i_{max}$
        Updated the particles $\chi_i^{P\text{-}best}$ and $\chi^{G\text{-}best}$ at iteration $t$
        **if** $C_{\Re}(\chi_i(t)) \leq C_{\Re}(\chi_i^{P\text{-}best}(t-1))$
            $\chi_i^{P\text{-}best}(t)=\chi_i(t)$
            ***else***
            $\chi_i^{P\text{-}best}(t)=\chi_i^{P\text{-}best}(t-1)$
        ***end (if)***

$$\chi^{G-best}(t) = \underset{1\leq i}{\arg\min}\, C_{\Re}(\chi_i^{P-best}(t))$$

        Update the state of the particle in the swarm using Eq.(17).
        Evaluate each candidate particle $\chi_i$ according to given cost function $C_{\Re}(\chi_i(t))$
    ***end (For)***
    Transfer best particles to next generation
***end (For)***
Output the corresponding route ($\chi^{G\text{-}best}$)

**Fig.3.** Pseudo-code of ACO, BBO, GA and PSO mechanism on TAMP approach

### *Application of BBO on Mission Planning Approach*

The population of habitat candidate solutions in BBO is initialized with feasible route sequence and each candidate solution holds a quantitative performance index representing the fitness of the solution called Habitat Suitability Index (HSI). High HSI solutions tend to share their useful information with poor HSI solutions. Habitability is related to some qualitative factors known as Suitability Index Variables (SIVs), which is a random vector of integers. Each candidate solution holds design variables of SIV, emigration rate (μ), immigration rate (λ) and fitness value of HSI. Generally, poor solution has higher immigration rate of λ and lower emigration rate of μ. Each given solution $h_i$ is modified according to probability of $P_s(t)$ that is the probability of existence of the $S$ species at time $t$ in habitat $h_i$ while $P_s(t+\Delta t)$ is the change in number of species after time $\Delta t$, given by Eq.(14).

$$\begin{aligned}
&P_S(t+\Delta t) = P_S(t)(1-\lambda_S\Delta t-\mu_S\Delta t) + P_{S-1}\lambda_{S-1}\Delta t + P_{S+1}\mu_{S+1}\Delta t \\
&\lambda_S = I*\left(1-\frac{S}{S_{\max}}\right);\quad \mu_S = E*\left(\frac{S}{S_{\max}}\right) \xrightarrow{\;if\;\; E=I\;} \lambda_S+\mu_S = E
\end{aligned} \tag{14}$$

where $I$ is the maximum immigration rate and $E$ is the maximum emigration rate. $S_{max}$ is the maximum number of species in a habitat. To have $S$ species at time $(t+\Delta t)$ in a specific habitat $h_i$, one of the following conditions must hold:

$$\begin{aligned}
&\forall h_i(t) \quad \exists\lambda_S, \mu_S, P_S(t) \\
&\begin{cases}
\forall h_i(t): \exists S \;\Rightarrow\; \forall h_i(t+\Delta t): \exists S \\
\forall h_i(t): \exists S \;\Rightarrow\; \forall h_i(t+\Delta t): \exists S-1 \\
\forall h_i(t): \exists S \;\Rightarrow\; \forall h_i(t+\Delta t): \exists S+1
\end{cases}
\end{aligned} \tag{15}$$

As suitability of habitat improves, the number of its species and emigration increases, while the immigration rate decreases. Afterward, the mutation operation is applied, which tends to increase the diversity of the population to propel the individuals toward global minima. Mutation is required for solution with low probability, while solution with high probability is less likely to mutate. Hence, the mutation rate $m(S)$ is inversely proportional to probability of the solution $P_s$.

$$m(S) = m_{\max}\left[\frac{1-P_S}{P_{\max}}\right] \tag{16}$$

where $m_{max}$ is the maximum mutation rate assigned by operator, $P_{max}$ is probability the habitat with maximum number of species $S_{max}$. The whole procedure of the BBO-based global route planning is summarized in Fig.3.

***Application of PSO on Mission Planning Approach***

The PSO start its process by initializing the particle population by feasible routes. Particle encoding is very important factor that affects effectiveness of the algorithm. Each particle involves a position and velocity that initialized randomly in a specific range in the search space and get updated iteratively according to Eq.(17) and then, the performance of particles is evaluated according to the defined cost functions. Each particle preserves it previous value, its best position $\chi^{P\text{-}best}$, and swarm global best position $\chi^{G\text{-}best}$. The current state value of the particle is compared to $\chi^{P\text{-}best}$ and $\chi^{G\text{-}best}$ at each iteration. Particle position and velocity get updated as follows.

$$\begin{aligned} \upsilon_i(t) &= \omega\upsilon_i(t-1) + c_1 r_1\left[\chi_i^{P-best}(t-1) - \chi_i(t-1)\right] + c_2 r_2\left[\chi_i^{G-best}(t-1) - \chi_i(t-1)\right] \\ \chi_i(t) &= \chi_i(t-1) + \upsilon_i(t) \end{aligned} \tag{17}$$

where $c_1$ and $c_2$ are acceleration coefficients, $\chi_i$ and $\upsilon_i$ are particle position and velocity at iteration $t$. $r_1$ and $r_2$ are two independent random numbers in [0,1]. $\omega$ exposes the inertia weight and balances the PSO algorithm between the local and global search. More detail about the algorithm can be found in (Kennedy and Eberhart 1995). The process of PSO is explained by Fig.3.

### 3.1.2 Path Planning Problem and Overview of the Applied Algorithms

***Application of the DE on OPP Approach***

The DE algorithm (Price and Storn 1997) is improved version of GA and uses similar operators of selection, crossover and mutation that is very suitable for synthetic natured problems like path planning (Zamuda and Sosa 2014, MahmoudZadeh et al. 2017-a). The DE produces better solutions and faster process due to use of real coding of floating point numbers in presenting problem parameters. The algorithm mostly relies on differential mutation and non-uniform crossover operations. Selection operator is applied then to converge the solutions toward the desirable regions in the search space. In first step, an initial population of solution vectors $\chi_i$, ($i=1,\ldots,$ $i_{\max}$) is randomly generated with uniform probability. For path planning purpose, any arbitrary path $\wp^i_{x,y,z}$ is assigned with solution vector $\chi_i^{x,y,z}$ where control points $\vartheta$ along the path $\wp^i_{x,y,z}$ correspond to elements of the solution vector. The solution space efficiently gets improved in each iteration $t$ applying evolution operators. A candidate solution vector is designated as

$$\begin{matrix} \forall i = 1,\ldots i_{\max} \\ \forall \chi_{i,t} \\ \forall t \in \{1,\ldots t_{\max}\} \end{matrix} \Rightarrow \exists \begin{cases} \chi_{i,t}^{x} = \vartheta_x^i(t) \\ \chi_{i,t}^{y} = \vartheta_y^i(t) \\ \chi_{i,t}^{z} = \vartheta_z^i(t) \end{cases} \tag{18}$$

where $i_{\max}$ is the number of individuals in DE population, $t_{\max}$ is the maximum number of iterations.

**[1]** ***Mutation***: The effectual modification of the mutation scheme is the main idea behind impressive performance of the DE algorithm, in which a weighted difference vector between two population members to a third one is added to mutation process that is called *donor*. Three different individuals of $\chi_{r1,t}$, $\chi_{r2,t}$, and $\chi_{r3,t}$ are selected randomly from the same iteration $t$, which one of this triplet is randomly selected as the *donor*. So, the mutant solution vector is produced by

$$\begin{aligned} &\dot{\chi}_{i,t} = \chi_{r3,t} + F(\chi_{r1,t} - \chi_{r2,t}) \\ &r1, r2, r3 \in \{1,\ldots i_{\max}\} \\ &r1 \neq r2 \neq r2 \neq i, \quad F \in [0,1+] \end{aligned} \tag{19}$$

where $F$ is a scaling factor that controls the amplification of the difference vector ($\chi_{r1,t} - \chi_{r2,t}$). Giving higher value to $F$ promotes the exploration capability of the algorithm. The proper donor accelerates convergence rate that in this approach is determined randomly with uniform distribution as follows

$$donor = \sum_{i=1}^{3}\left(\lambda_i \bigg/ \sum_{j=1}^{3}\lambda_j\right)\chi_{ri,t}, \tag{20}$$

where $\lambda_j$ ∈[0,1] is a uniformly distributed value. This scheme provides a better distribution of the solution vectors. The mutant individual $\dot{\chi}_{i,t}$ and parent individual $\chi_{i,t}$ are then shifted to the crossover operation.

**[2]** ***Crossover*****:** The parent vector to this operator is a mixture of individual $\chi_{i,t}$ from the initial population and the mutant individual $\dot{\chi}_{i,t}$. The produced offspring $\ddot{\chi}_{i,t}$ from the crossover is described by

$$\begin{cases} \chi_{i,t} = (x_{1,i,t},\ldots,x_{n,i,t}) \\ \dot{\chi}_{i,t} = (\dot{x}_{1,i,t},\ldots,\dot{x}_{n,i,t}) \\ \ddot{\chi}_{i,t} = (\ddot{x}_{1,i,t},\ldots,\ddot{x}_{n,i,t}) \end{cases} \Rightarrow \ddot{x}_{j,i,t} = \begin{cases} \dot{x}_{j,i,t} & rand_j \leq r_C \vee j = k \\ x_{j,i,t} & rand_j \leq r_C \wedge j \neq k \end{cases} \quad j = 1,\ldots n; \quad n \in [1, i_{\max}] \tag{21}$$

where $k$ ∈{1,…, $i_{\max}$}is a random index chosen once for all population $i_{\max}$. The second DE control parameter is the rate of crossover $r_C$ ∈ [0,1] that is set by the user.

**[3]** ***Evaluation and Selection*****:** The offspring produced by the crossover and mutation operations is evaluated. The best-fitted solutions are selected and transferred to the next generation ($t+1$).

$$\dot{\chi}_{i,t+1}=\begin{cases}\dot{\chi}_{i,t} & C_{\wp}(\dot{\chi}_{i,t})\leq C_{\wp}(\chi_{i,t})\\ \chi_{i,t} & C_{\wp}(\dot{\chi}_{i,t})>C_{\wp}(\chi_{i,t})\end{cases};\quad \ddot{\chi}_{i,t+1}=\begin{cases}\ddot{\chi}_{i,t} & C_{\wp}(\ddot{\chi}_{i,t})\leq C_{\wp}(\chi_{i,t})\\ \chi_{i,t} & C_{\wp}(\ddot{\chi}_{i,t})>C_{\wp}(\chi_{i,t})\end{cases} \tag{22}$$

The efficiency of the offspring and parents are compared for each operator by defined cost function $C_{\wp}$ and the worst individuals are eliminated from the population. The process of DE is presented in Fig.5.

### *Application of the FA on OPP Approach*

The FA is another swarm-intelligence-based meta-heuristic algorithm inspired from the flashing patterns of fireflies, in which the fireflies attracted to each other based on brightness and regardless of their sex (Yang 2009). As the distance of fireflies increases their brightness get dimmed. The less bright firefly approaches to the brighter one. Attraction of each firefly is proportional to its brightness intensity received by adjacent fireflies and their distance $L$. The attraction factor $\beta$ and movement of a firefly $\chi_i$ toward the brighter firefly $\chi_j$ is calculated by

$$\begin{aligned}&\beta=\beta_0 e^{-\varepsilon L^2}\\ &\chi_i^{t+1}=\chi_i^t+\beta_0 e^{-\varepsilon {L_{ij}}^2}(\chi_j^t-\chi_i^t)+\alpha_t\varsigma_i^t\\ &\alpha_t=\alpha_0\kappa^t,\quad \kappa\in(0,1)\end{aligned} \tag{23}$$

where $\beta_0$ is the attraction value at $L$=0, $\alpha_t$ is the randomization parameter that control the randomness of the movement and can be tuned iteratively. The $\alpha_0$ is the initial randomness scaling value and $\kappa$ is the damping factor. The $\varsigma_i^t$ is a random vector generated by uniform distribution Gaussian distribution at time $t$. There should be a proper balance between engaged parameters, because if the $\beta_0$ approaches to zero, the movement turns to a simple random walk, while ε= 0 turns it to a variant of particle swarm optimization (Yang 2009). To the purpose of path planning the fireflies assigned with path control points $\chi_i=\vartheta_i^{x,y,z}$ where $\chi_1=\vartheta_1^{x,y,z}$ and $\chi_n=\vartheta_n^{x,y,z}$ corresponds to coordinates of the starting and ending control points. Afterward, the generated path that coded by fireflies tends to be optimized iteratively toward the best solution in search space in accordance with defined cost function $C_{\wp}$. Two major advantages make the FA more efficient comparing to other optimization algorithms. First, the FA naturally uses an automatic subdivision approach that promotes its capabilities in dealing with multimodality. The attraction decreases with distance and in this context entire population automatically get subdivided, where each subgroup swarm around its local optimum and the best global solution is selected among the set of the local optimums. This fact increases convergence rate of the algorithm. Second, in a case that population is sufficiently larger than the number of modes, such an automatic subdivision nature motivates fireflies to find all optima iteratively and simultaneously, which makes the FA specifically suitable and flexible in dealing with continuous problems, highly nonlinear problems, and multi-objective problems (Yang and He 2013). The control parameters in FA can be tuned iteratively that increases convergence rate of the algorithm too. The path encoding and pseudo-code of FA process on path planning approach is provided by Fig.4 and Fig.5, respectively.

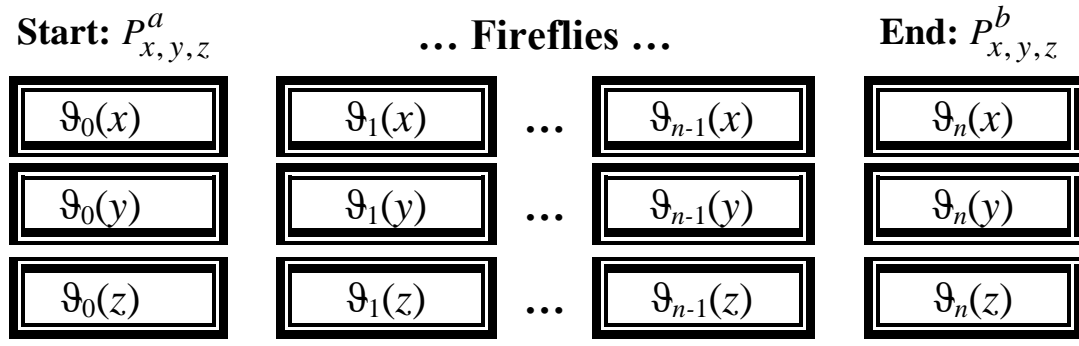

**Fig.4.** Encoding scheme of a fireflies by path control points

where $\vartheta_0(x,y,z)$ and $\vartheta_n(x,y,z)$ corresponds to coordinates of the starting and ending points, $\vartheta_i(x,y,z)$, is any arbitrary control point along the generated path that coded by fireflies and tend to be optimized toward the best solution in search space in accordance with defined optimization indexes an constraints as the FA algorithm iterates. The path curve is reconstructed based on the proposed encoding scheme and Eq.(23) in which all points should be located in the feasible area otherwise the path gets violation.

### *Application of the BBO on OPP Approach*

A general overview of the BBO mechanism is provided in Sections 3.1.1, so we directly turn to BBO mechanism on path planning procedure. In the proposed path planner, each habitat $h_i$ of BBO corresponds to the coordinates of the B-spline control point's $\vartheta_i$ that utilizes in path generation, where $h_i$ defined as a parameter to be optimized ($P_s$:($h_1$,$h_2$,…,$h_{n-1}$)). As the BBO algorithm iterates, each individual habitat gets attracted toward its respective best position based on habitat suitability index variable (SIV). The emigration and immigration rate tend to improve the solutions. The HSI value indicates the objective function value for a candidate solution that is intended to be maximized by the algorithm according to solution's emigrating and immigrating features. Each solution of the population should be evaluated before starting the optimization process. A poor solution has higher immigration rate of λ and lower emigration rate of μ. The immigration rate λ is used to probabilistically modify the SIV of a selected solution $h_i$. Then emigration rates μ of the other solutions are considered and one of them probabilistically selected to migrate its SIV to solution $h_i$. Each given solution $h_i$ is modified according to probability of $P_s(t)$ that is the probability of existence of the $S$ species at time $t$ in habitat $h_i$. The pseudo code of the BBO algorithm and its mechanism on path planning process is provided in Fig.5.

### *Application of the PSO on OPP Approach*

The argument for using PSO in path planning problem is strong enough due to its superior capability in scaling well with complex and multi-objective problems. The PSO operates in a continuous space originally and it is well suited for solving vehicles path planning problem due to continuous nature of this problem. A general overview of the PSO algorithm is provided in Section 3.1.1 and its mechanism on path planning process is provided by following pseudo code in Fig.5. Each particle in the swarm assigned by a potential path. The position and velocity parameters of the particles corresponds to the coordinates of the b-spline control points $\vartheta_i$. As the PSO algorithm iterates, every particle is attracted towards its respective local attractor based on the outcome of the particles individual search as well as the particles swarm search results.

**Procedure of FA on OPP**

Initialization phase:

- Initialize population of fireflies $\chi_i$ with the control points $\vartheta_i$
- Define light absorption coefficients ε
- Initialize the attraction coefficient $\beta_0$
- Set the damping factor of $\kappa$
- Initialize the randomness scaling factor of $\alpha_0$
- Set the parameter of randomization $\alpha_t$
- Set the maximum iteration $t_{max}$
- Set the number of population $i_{max}$

***For*** $t$ =1 ***to*** $t_{max}$
***For*** $i$ =1 ***to*** $i_{max}$
Reconstruct a path according to $\chi_i$
Evaluate the path $C_{\wp}(\chi_i(t))$
Update light intensity of $\chi_i$
***For*** $j$ =1 ***to*** $i$
Reconstruct a path according to $\chi_j$
Evaluate the path $C_{\wp}(\chi_j(t))$
Update light intensity of $\chi_j$
***If*** ($\beta j > \beta i$),
Move firefly $i$ towards $j$
***end (if)***
***end (For)***
***end (For)***
Rank the fireflies and find the current best
***end (For)***
Output result

**Procedure of DE on OPP**

Initialization phase:

- Initialize population of solution vectors randomly $\chi_i^{x,y,z}$ with the control points $(\vartheta^i_x, \vartheta^i_y, \vartheta^i_z)$
- Set the maximum number of iteration $t_{max}$
- Choose appropriate parameters for the population size $i_{max}$
- Set the mutation coefficient
- Set the crossover coefficient

***For*** $t$ =1 ***to*** $t_{max}$
***For*** $i$ =1 ***to*** $i_{max}$
Reconstruct a path according to $\chi_{i,t}$
Evaluate the path
Determine the *donor*
Apply mutation using
$\dot{\chi}_{i,t} = \chi_{r3,t} + F(\chi_{r1,t} - \chi_{r2,t})$
$r1, r2, r3 \in \{1,\dots,i_{\max}\}, r1 \neq r2 \neq r2 \neq i, \quad F \in [0,1+]$
***For*** $j$ =1 ***to*** $i$
Apply crossover using $\chi_{i,t}$ mutant solution $\chi^{\cdot}_{i,t}$ to get the $\chi^{\cdot\cdot}_{i,t}$
Reconstruct a path according to $\chi_{i,t}$, $\chi^{\cdot}_{i,t}$, and $\chi^{\cdot\cdot}_{i,t}$
Evaluate the corresponding paths to $\chi_{i,t}$, $\chi^{\cdot}_{i,t}$, and $\chi^{\cdot\cdot}_{i,t}$
***if*** $C_{\wp}(\chi_{i,t}) \leq C_{\wp}(\chi^{\cdot}_{i,t})$
$\chi^{\cdot}_{i,t+1} = \chi_{i,t}$
***if*** $C_{\wp}(\chi_{i,t}) \leq C_{\wp}(\chi^{\cdot\cdot}_{i,t})$
$\chi^{\cdot}_{i,t+1} = \chi_{i,t}$
***else***
$\chi^{\cdot}_{i,t+1} = \chi^{\cdot\cdot}_{i,t}$
***end (if)***
***else***
$\chi^{\cdot}_{i,t+1} = \chi^{\cdot}_{i,t}$
***end (if)***
***end (For)***
***end (For)***
Select the best solutions to transfer to next iteration
***end (For)***
Output best solution and the corresponding paths

**Procedure of BBO on OPP**

Initialize a set of solutions as initial habitat population

- Assign B-spline control points $\vartheta_i$ as habitat $h_i$
- Choose appropriate parameters for the population size $i_{max}$
- Set the number of control-points ($n$) that used to generate the B-Spline path
- Set the maximum number of iteration $t_{max}$
- Assign maximum immigration and emigration rate ($I$, $E$)
- Assign maximum mutation rate $m(S)$
- Set $S_{max}$ and SIV vector

***For*** $t$=1 ***to*** $t_{max}$
Compute immigration rates λ and emigration rate μ for each solution
$\lambda_S(t) = I * \left(1 - \frac{S}{S_{\max}}\right); \quad \mu_S(t) = E * \left(\frac{S}{S_{\max}}\right)$
Evaluate the fitness (HSI) of each habitat and identify Elite Habitats based on HIS
Modify habitats based on λ and μ (Migration):
$P_S(t) = P_S(t-1)(1 - \lambda_S(t) - \mu_S(t)) + P_{S-1}\lambda_{S-1}(t) + P_{S+1}\mu_{S+1}(t)$
***For*** $i$=1 ***to*** $i_{max}$
Use $\lambda_i$ to probabilistically decide whether immigrate to habitat $h_i$
***if*** $rand(0,1) < \lambda_i$
***For*** $j$=1 ***to*** $i_{max}$
Select the emigrating habitat $h_j$ with probability $\propto \mu_j$
**if** $rand(0,1) < \mu_j$
Replace a randomly selected SIV variable of $h_i$ with its corresponding value in $h_j$
***end (if)***
***end (For)***
***end (if)***
***end (For)***
Carry out the mutation based on probability by $m(S) = m_{\max}\left[\frac{1 - P_S}{P_{\max}}\right]$
Transfer the best solution in the population from one generation to the next
***end (For)***
Output the best habitat and its correlated path as the optimal solution

**Procedure of PSO on OPP**

Initialize each particle by random velocity and position in following steps:

- Assign B-Spline control points $\vartheta_i$ as particle position $\chi_i$
- Initialize each particle with random velocity $\upsilon_i$ in range of predefined bounds $\beta^i_{\vartheta} = [U^i_{\vartheta}, L^i_{\vartheta}]$.
- Choose appropriate parameters for the population size $i_{max}$
- Set the number of control-points $m$ used to generate the B-Spline path
- Set the maximum number of iterations $t_{max}$
- Initialize $\chi_i^{P\text{-}best}(1)$ with current position of each particle at first iteration $t$=1.
- Set the $\chi^{G\text{-}best}(1)$ with the best particle in initial population at $t$=1.

***For*** $t$=1 ***to*** $t_{max}$
Evaluate each candidate particle according to given cost function
***For*** $i$=1 ***to*** $i_{max}$
Updated the particles $\chi_i^{P\text{-}best}$ and $\chi^{G\text{-}best}$ at iteration $t$
**if** $C_{\wp}(\chi_i(t)) \leq C_{\wp}(\chi_i^{P\text{-}best}(t\text{-}1))$
$\chi_i^{P\text{-}best}(t) = \chi_i(t)$
***else***
$\chi_i^{P\text{-}best}(t) = \chi_i^{P\text{-}best}(t\text{-}1)$
***end (if)***
$\chi^{G-best}(t) = \underset{1 \leq i}{\arg\min}\, C_{\wp}(\chi_i^{P-best}(t))$
Update the state of the particle in the swarm
$\upsilon_i(t) = \omega\upsilon_i(t-1) + c_1 r_1\left[\chi_i^{P-best}(t-1) - \chi_i(t-1)\right] + c_2 r_2\left[\chi_i^{G-best}(t-1) - \chi_i(t-1)\right]$
$\chi_i(t) = \chi_i(t-1) + \upsilon_i(t)$
Evaluate each candidate particle $\chi_i$ according to given cost function $C_{\wp}(\chi_i(t))$
***end (For)***
Transfer best particles to next generation
***end (For)***
Output $\chi^{G\text{-}best}$ and its correlated path as the optimal solution

**Fig.5.** Pseudo-code of DE, FA, BBO and PSO mechanism on path planning problem

## 4 Performance Evaluation of $R^{II}MP^{II}$

The model aims to use the total time for completing the best possible tasks in the most efficient way and guarantee on-time arrival to the destination while concurrently check for safe deployment through an efficient path planning strategy. To this purpose an accurate and stringent concurrency between different components of the model is required so that each component can carry out its responsibility in a synchronous manner. Various population based meta-heuristic approaches are tested in simulation and the stability and real-time applicability of the proposed $R^{II}MP^{II}$ framework. A quantitative analyze is provided in order to examine the capability and efficiency of the proposed framework performed in MATLAB®2016. This study aims to prove efficiency of meta-heuristics against any other deterministic methods for the purpose of this research rather than criticizing or comparing efficiency of the applied algorithms. This implementation includes the robustness assessment of the framework in diverse experiments in terms of mission timing, multitasking, safe and efficient deployment to ensure its reliability and robustness regardless of the applied algorithms. Accuracy and performance of all different combination of the employed algorithms has been studied globally in different situations that closely matches the real-world conditions.

## 4.1 Analysis of TAMP Performance

Comparison of different Meta heuristic methods is common due to their similarity in optimization mechanism. Any of these algorithms show different performance on different problems; however, apparently all of them are capable of producing feasible solution after the first iteration has completed. Second benefit of these algorithms is their ability of handling non-linear cost functions. As mentioned earlier, obtaining a near optimal solution in faster computation time is more important for the purpose of this research where the modelled environment corresponds to a spatiotemporal time varying terrain with high uncertainty. The real-time performance of the applied algorithms on proceeding problem of TAMP is not fully known; hence the computation time would be one of the important performance indicators for evaluating the applied methods. With respect to the defined cost function for the TAMP problem (given by Eq.(7)), an optimum solution corresponds to a mission that takes maximum use of available time by maximizing number of completed tasks and route weight, while respects the upper bound time threshold. Accordingly, performance of the ACO, BBO, GA, and PSO algorithms is investigated and compared in a quantitative manner through 150 runs of Monte Carlo simulation, given by Fig.6.

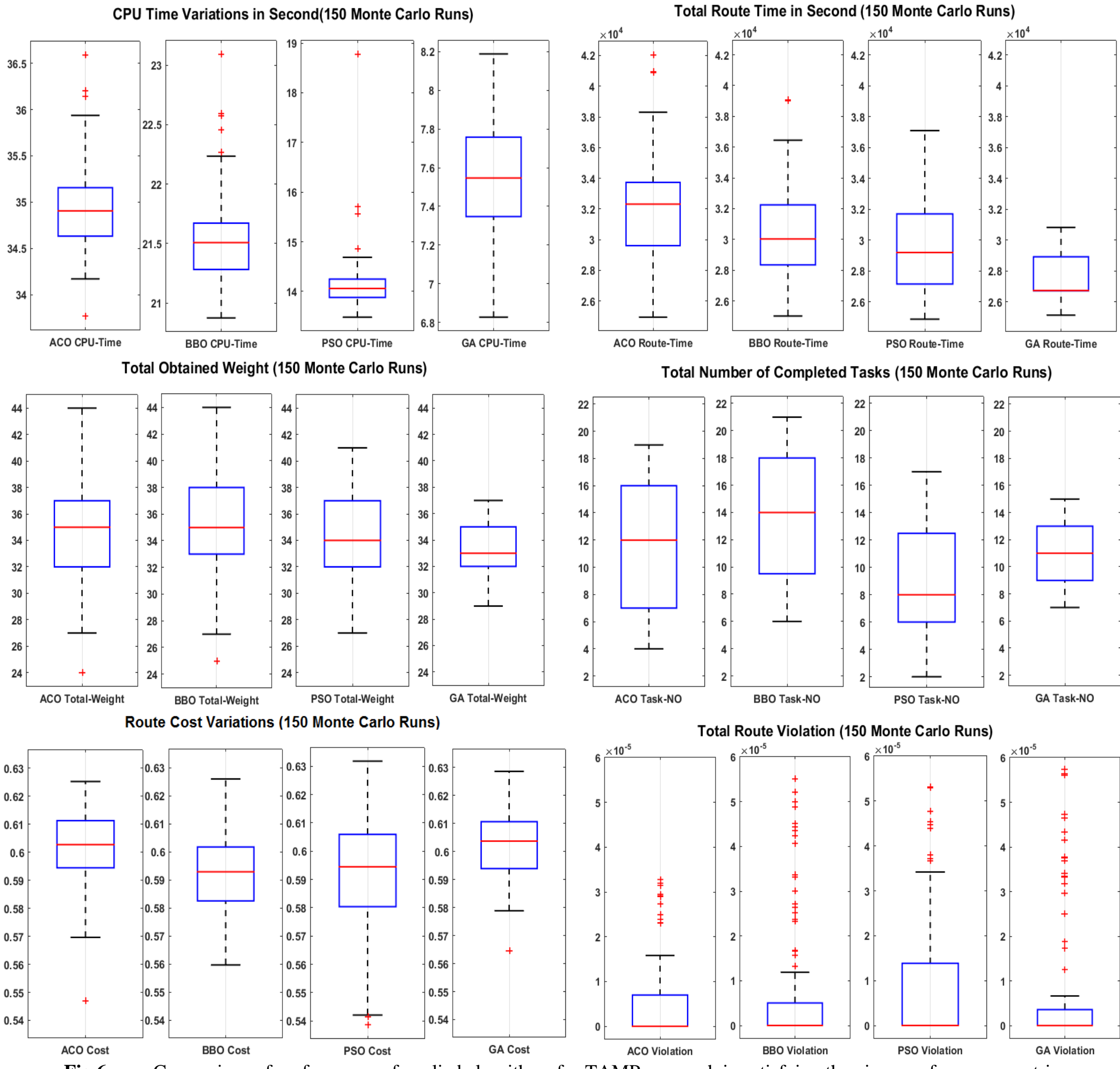

**Fig.6.** Comparison of performance of applied algorithms for TAMP approach in satisfying the given performance metrics

In the performed simulation, number of vertices of the graph is set to differ between 30 to 50 nodes, connection between nodes and overall graph topology is randomly changed (using a Gaussian distribution) on the problem search space in each execution. The time threshold is set on $3.42{\times}10^4$ (*sec*). Fig.6 compares the functionality of the applied algorithms in dealing with problem's space deformation. It is apparent from CPU-Time variations that all algorithms are capable of satisfying real-time requirement of the proposed problem as the variation for all algorithms is drawn in a very narrow boundary in range of seconds; however, the fastest operation belongs to GA and then PSO, BBO, and ACO, respectively. The appropriate case for the TAMP approach is producing routes with maximum travel time limited to threshold. It is denoted from route time variations in Fig.6, all algorithms significantly manage the route time to approach proposed time threshold, but it seems the GA acts more cautious in constraining the route time as the variation range for GA is strictly kept far below the defined threshold. However, variations of PSO, BBO, and ACO almost placed in a same range, but no outlier is seen in PSO route time variations, which is good point for this algorithm.

Besides, stability of TAMP model is demonstrated in Fig.6 based on outcomes of each algorithm in terms of completed tasks and total obtained weight. It is apparent that variation range for total obtained weight and completed tasks by GA appeared in a smaller range in comparison with outcome of others respecting the narrow range of GA route time. It seems that three other algorithms of BBO, ACO and PSO propose almost similar performance in quantitative measurement of two significant mission's metrics of route obtained weight and number of covered tasks; however, being more specific BBO acts more efficiently as the produced results by BBO suppresses two others. it is noteworthy to mention that the route costs is varying in a similar range for all four algorithms, almost between 0.57 to 0.62, and it is noted that average variation of route violation for Monte Carlo executions approaches zero, which confirms efficiency of the applied algorithms in satisfying defined constraints for TAMP and they are persistent against problem space deformation, as this is a challenging problem for other deterministic algorithms addressed earlier. Being more specific, the provided results also shows that the GA reveals better performance in eliminating the route violation. To compare applied metaheuristic methods, a benchmark table is employed as presented in Table.1 that gives a summary of the characteristics of four aforementioned algorithms quantitatively analyzed by Fig.6.

Table 1. Comparison of the GA, BBO, PSO, and ACO based mission planning performance

| | *CPU Time* | *Route Time* | *Route Weight* | *Covered Tasks* | *Route Cost* | *Route Violation* |
|---|---|---|---|---|---|---|
| ***Best to Worst*** | GA | BBO | BBO | BBO | PSO | GA |
| | PSO | PSO | ACO | ACO | BBO | BBO |
| | BBO | ACO | PSO | PSO | ACO | ACO |
| | ACO | GA | GA | GA | GA | PSO |

The ACO and PSO has a similar feature in which both use three principals of the transient state, local and global updates. Local updating in ACO preserves the quality of the population in each iteration so it requires less iterations to converge the population toward the optimum solution. However, both PSO and ACO face the stagnation problem when the swarm quickly gets stuck in a local optimum; hence, the algorithm is not able to promote solution quality as time progresses. The particle or ant swarm will eventually find an optimal or near optimal solution despite no estimate can be given on convergence time. The ACO also shares the similar shortcoming with the GA algorithm as there is no guarantee in random method for fast optimal solutions even though GA select a proper population at each iteration. Moreover, when complexity of graph increases the length of chromosomes or solution vectors is increased accordingly, so that mission planning process becomes slower; nevertheless, all aforementioned algorithms operate in a competitive CPU time unlike to deterministic and heuristic methods. Hence, the evolutionary algorithms are suitable to produce optimal solutions quickly for real-time applications. Compared performance of four applied algorithms in routing and mission planning process declares the superior performance of BBO comparing others; however, there is only a slight difference between the produced results.

### 4.2 Analysis of OPP Performance

The efficient path planner is able to make a use of desirable currents and cope undesirable current flows to increase battery life time. Besides, the uncertainty of operating filed, which more or less is taken into account in the aforementioned literature, have a devastating effect vehicles deployment and safety. According to define path cost function $C_{\wp}$, the optimum solution corresponds to quickest and safest path in an uncertain terrain underwater environment. Performance of the path planner in coping current variations and collision avoidance, therefore, need to be addressed thoroughly in accordance with complexity of the modelled terrain. The path planning in current study mainly focuses on finding the optimum trajectory adjustments using the advantages of desirable current field while avoid colliding obstacles. Current research takes the advantages of the previous research (MahmoudZadeh et al. 2016-c) in modelling uncertain static and dynamic obstacles, in which obstacles are generated randomly and configured individually with random position and velocity. The obstacles position uncertainty assumed to grow linearly with time at rates randomly selected based on type of the obstacle with or without encountering current velocity. Any time that path planner is recalled the new optimal path is generated from current position to the destination by refining the previous solution to cope the changes of the terrain. The terrain updates in this simulation involves update of 2D static current map (two static current map are conducted) and obstacles deformation over time. These updates regularly measured from the on-board sonar sensors and according to their present state, their behaviour in one step forward can be estimated during the next execution period. Fig.7 performs the path curves produced by introduced evolution-based OPP for DE, FA, PSO, BBO algorithms in adapting ocean currents and handling obstacle avoidance. The current field in Fig.7 is computed from a random distribution of 5 to 10 vortices in 3.5 $km^2$ spatial domain. The updated current map (presented by black arrows) contain similar features of the previous map (given by blue arrows). The OPP in the proceeding research simultaneously tracks the measurements of the terrain status and tends to generate the shortest collision-free trajectory from the starting position denoted by "○" to the destination point marked by "□". No collision will occur if the trajectory does not cross inside the collision boundaries of obstacles or coastal areas on the map. The collision boundaries represented as black circles around the obstacles indicating a confidence of 98% that the obstacle is located within this area. The gradual increment of collision boundary of each obstacle is presented in Fig.7(d) in which the uncertainty propagation is assumed linear with time.

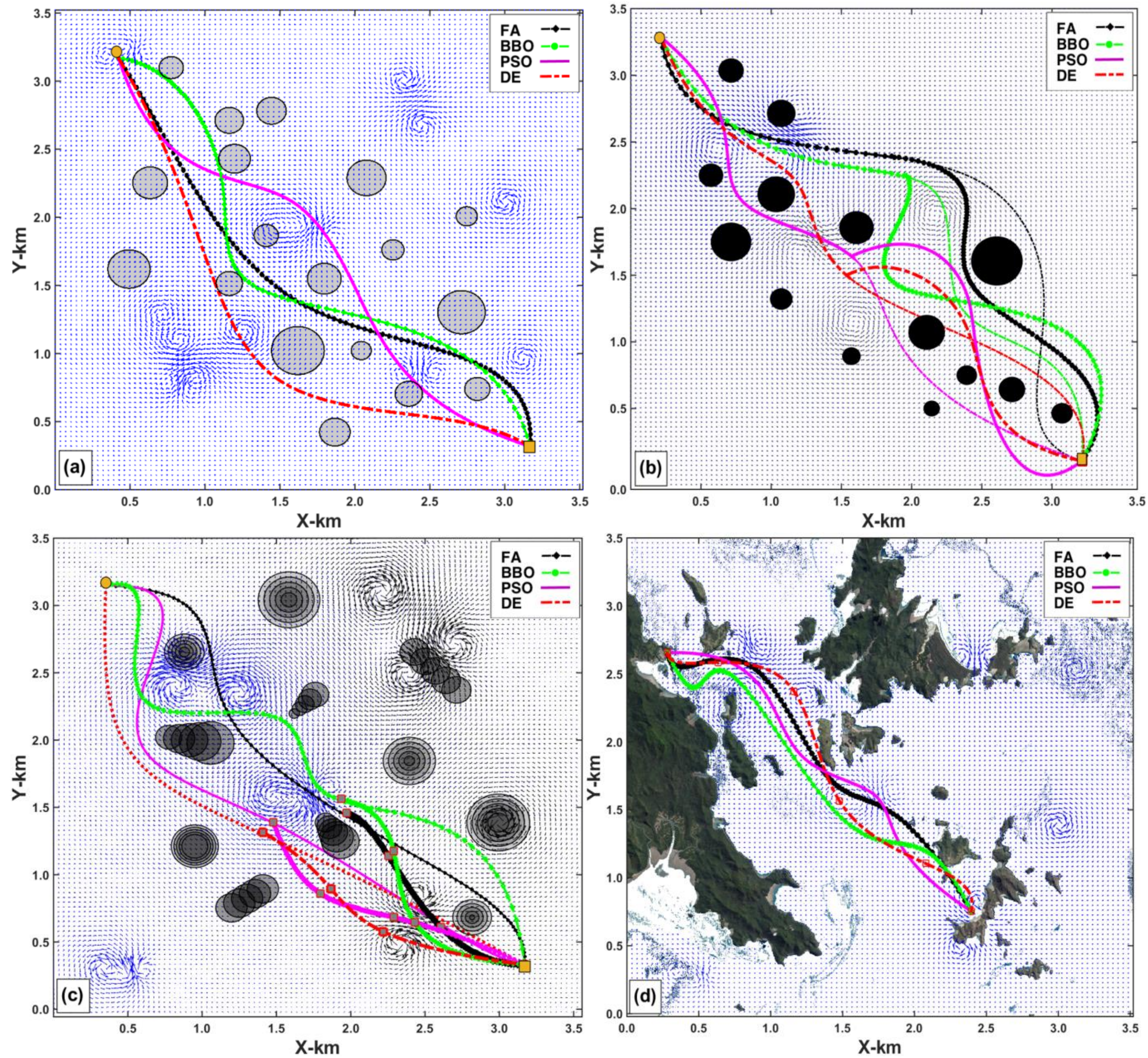

**Fig.7**. **(a)** Evolution curves of FA, DE, BBO, and PSO algorithms in coping current flow and obstacles avoidance encountering one static current map and random number of static obstacles; **(b)** The path behaviour and re-planning process to update of the water current; **(c)** Path curves behaviour and its deformation in handling collision avoidance according to current map updates encountering static/moving uncertain obstacles; **(d)** Accuracy of the proposed path planners in recognizing the map coastal sections encountering water current effects.

Referring to Fig.7(a), it is apparent that the initial produced path by all algorithms adapts the behaviour of current and accurately handles obstacle avoidance. Being more specific, it seems that the FA shows better performance in adapting adverse or concordant current arrows and take shortest path comparing to others. As presented in Fig.7(b-c), the proposed OPP is capable of re-generating the alternative trajectory according to latest update of current map, in which the regenerated path takes a detour taking the advantages of previously generated optimum path data and the desirable current flow to speed up and save more energy in its motion toward the destination. Further, it is noted from Fig.7(b,c) the path generated by FA shows its superior flexibility in coping with current change specifically when the current magnitude gets sharper. Increasing the number of obstacles and adding uncertainty into account, increases the problems complexity; however, it is derived from Fig.7(c) all aforementioned algorithms are able to efficiently carry out the collision avoidance operation regardless of terrain complexity. As presented by Fig.7(b-c) path re-planning procedure reuses the history of previous paths (online re-planning) to achieve more optimized path, where the initial paths are presented by thinner lines and new paths are shown by thicker lines.

Fig.7(d) compares performance of the four algorithms in recognizing coastal areas of the map as forbidden zones for deployment along with adding information of static current map into account. In the implementation of the k-means clustering method, the given map is classified to 3 subsections of uncertain risky area, coastal area, and water covered area as valid zone for deployment. It is derived from simulation results in Fig.7(d) that there is only slight difference comparing the results produced by all four algorithms and all of them accurately avoid crossing into the coastal sections of map even when the disturbing current pushes the vehicle to the undesired directions, also they are accurate and resistant against current deformations in the given operation window. For making a better comparison between performances of the applied algorithms in OPP, their cost and violation variations (given by Eq.(10)) is investigated and presented through the Fig.8 and Fig.9.

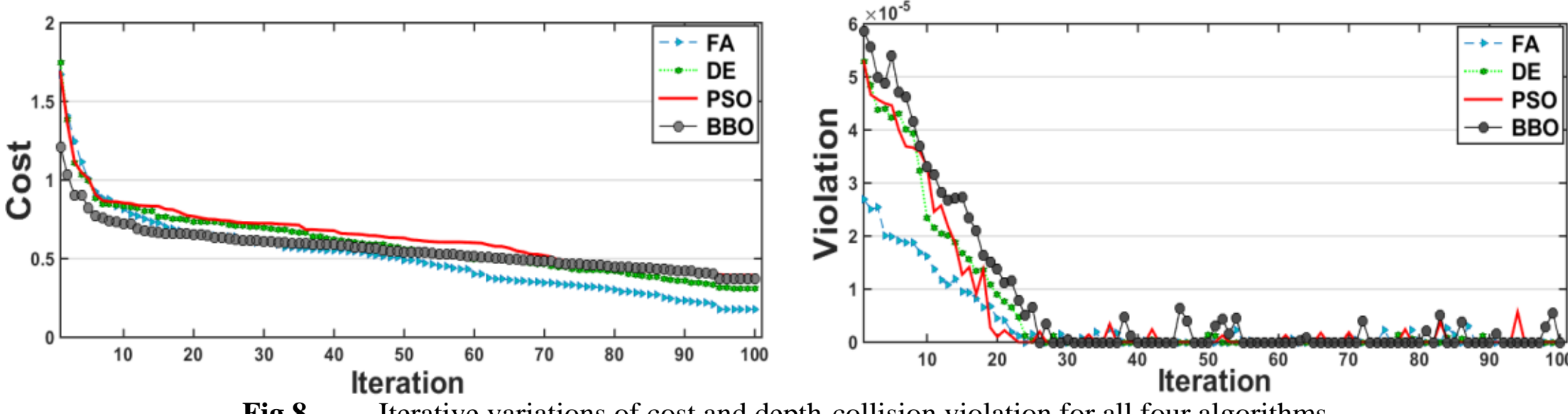

**Fig.8.** Iterative variations of cost and depth-collision violation for all four algorithms.

The cost variations of all four algorithms given by Fig.8 declares that the FA shows superior accuracy in reducing the cost within 100 iterations while the DE and PSO do this process almost in a same accuracy; after all, the BBO do this process with lowest accuracy despite its starting point even stands in a lower value $C_{\wp}\approx 1.2$ comparing others. It is also noted from variation of the depth-collision violation in Fig.8 the value of violation gradually diminishes iteratively for all four algorithm, but again the FA shows the best performance in managing solutions toward eliminating the violation. On the other hand, the B-spline paths' curve is obtainable by the vehicle's radial acceleration and angular velocity constraints. The behavior of the produced path curves by FA, DE, BBO, and PSO algorithms in satisfying the vehicular constraints is presented through the Fig.9.

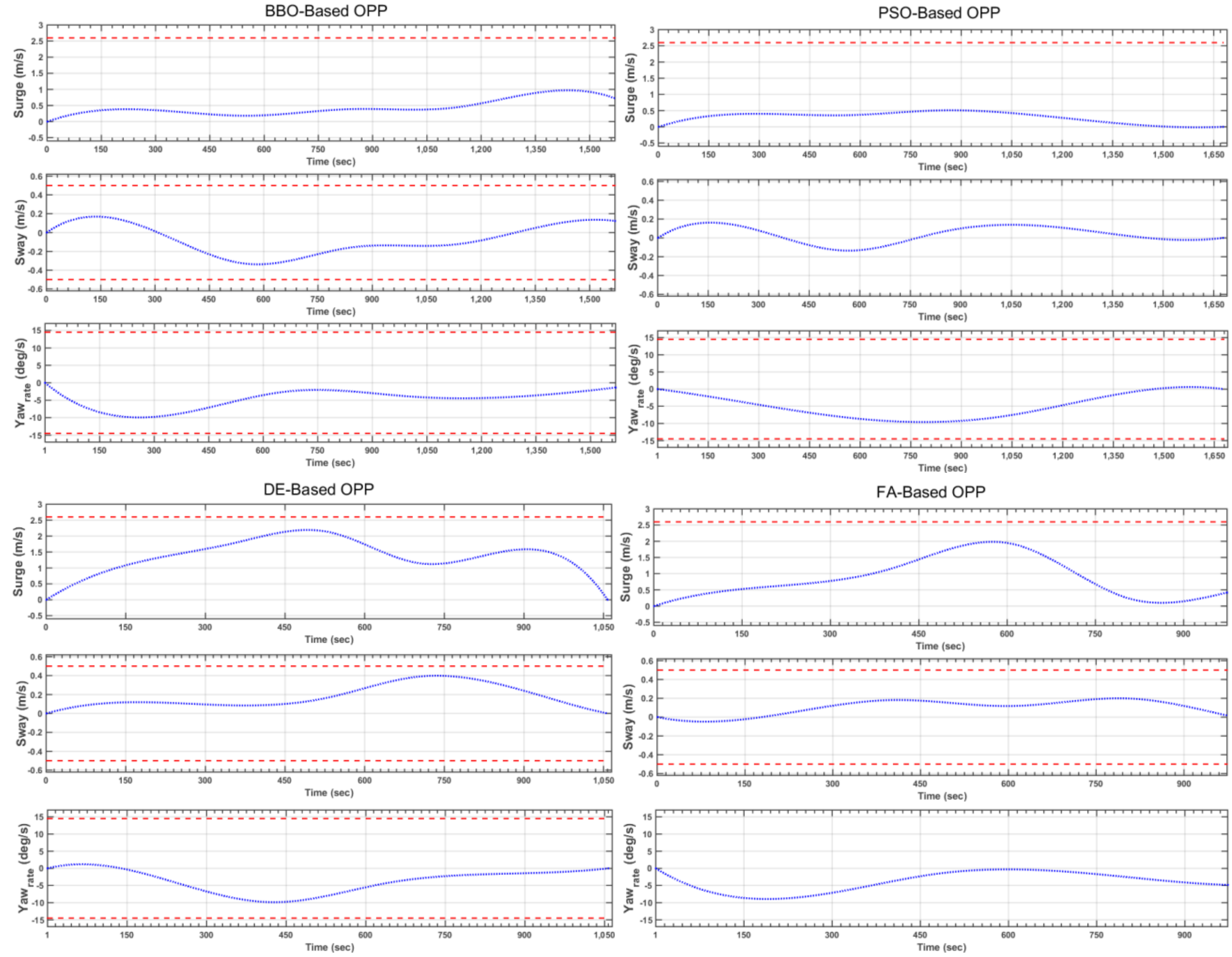

**Fig.9.** Variations of the vehicular constraint of Surge, Sway, and Yaw rate over the path flight time

Considering the variations of surge, sway, and yaw rate of produced paths by FA, BBO, DE, and PSO in Fig.9, it is evident that all generated paths accurately satisfy the defined constraints as the produced lines are bounded to specified violation boundaries (presented with the red dashed line). It is outstanding from the simulation results given by Fig.7, Fig.8 and Fig.9, all FA, DE, BBO and PSO based path planners perform almost similar cost and great fitness encountering environmental and vehicular constraints. Being more specific, it is notable that FA acts more efficiently in finding time optimum current-resistant collision-free solutions and performs faster convergence with lower cost comparing to other algorithms. Albeit, it is also noteworthy to mention that performance and optimality of all generated paths by four algorithms are competitive and very similar to each other.

### 4.3 $R^{II}MP^{II}$ Robustness Analysis

To measure the performance optimality of the model, several performance metrics are utilized such as, total model cost and violation that involves the resultant violation of both OPP and TAMP operations; mission profit which belongs to number and total value of the completed tasks in a single mission; real-time performance of the system; and mission timing. The aforementioned factors are analyzed in a single run for a different combination of the applied algorithms for both OPP and TAMP, given by Fig.10 and Fig.11, and then 150 runs of Monte Carlo simulation is carried out to prove the robustness of the introduced model employing any meta-heuristic method (presented by Fig.12 to Fig.15), which is a remarkable achievement for a real time autonomous system.

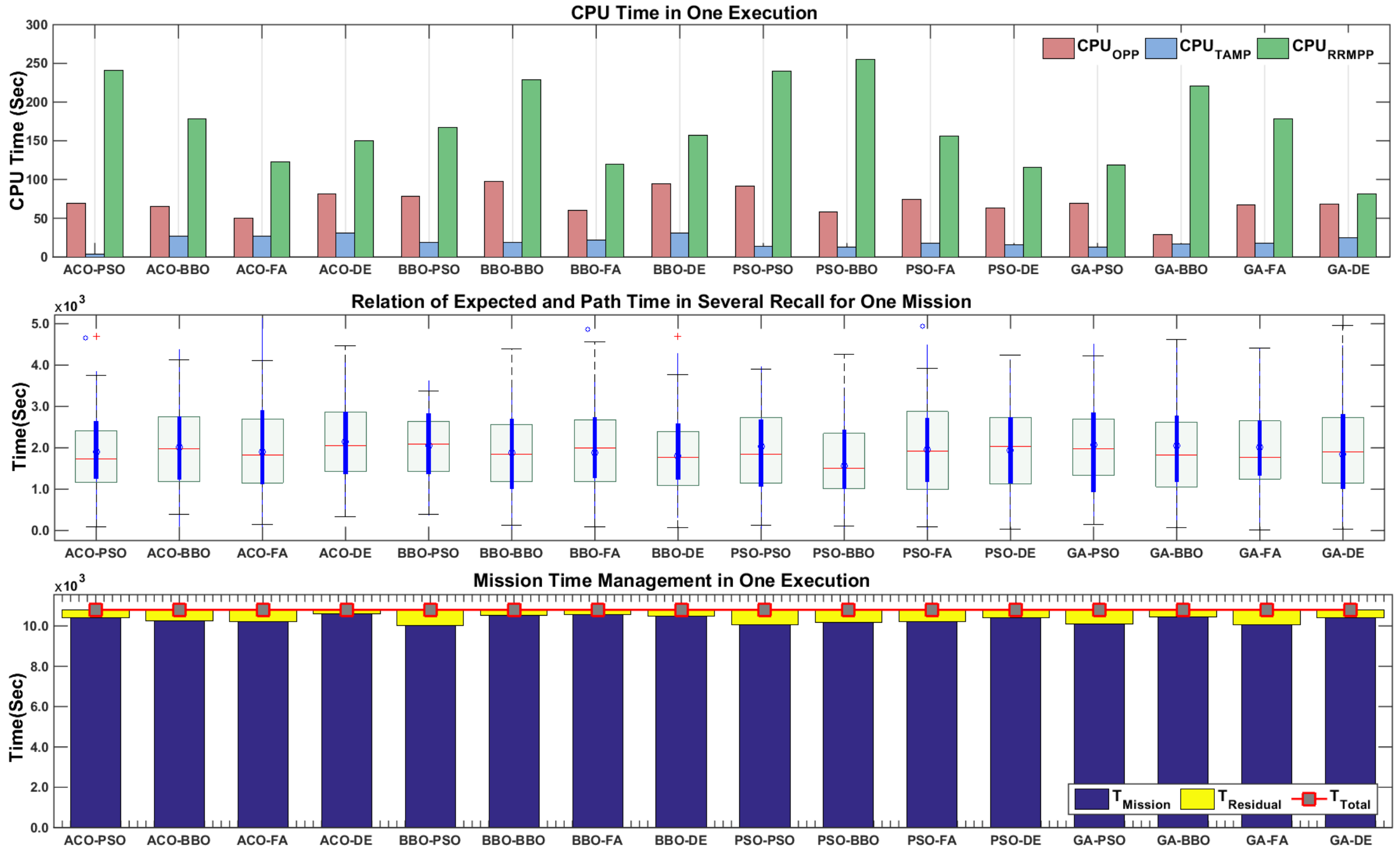

**Fig.10.** **(a)** The computational time of the OPP, TAMP modules and entire $R^{II}MP^{II}$ operation in one execution run for different combination of aforementioned algorithms; **(b)** Concurrency of the TAMP and OPP based on consistency of the expected and path flight time; **(c)** Mission timing performance of the $R^{II}MP^{II}$ model.

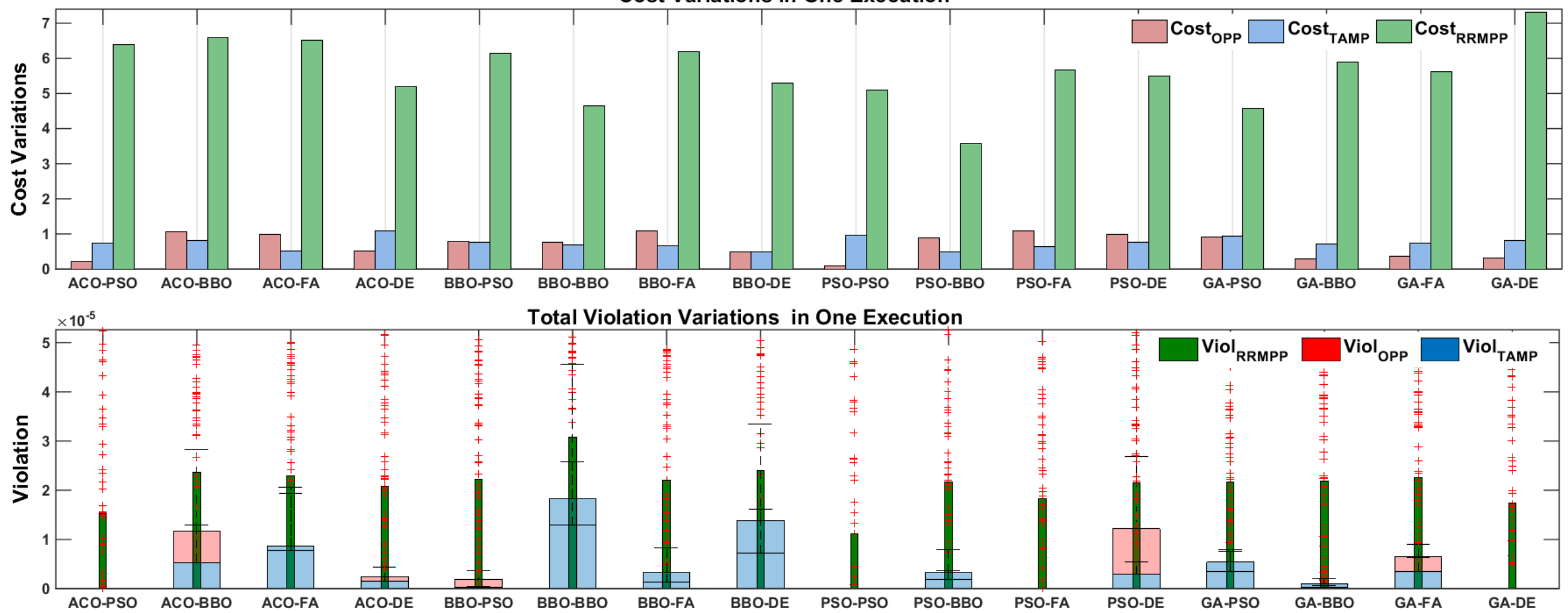

**Fig.11.** The cost and violation variation of the OPP, TAMP modules and entire $R^{II}MP^{II}$ operation in one execution run for different combination of aforementioned algorithms.

The most critical factor for preserving the robustness and consistency of the proposed real-time $R^{II}MP^{II}$ scheme is maintaining comparably fast operation for each component of the system (TAMP-OPP) to prevent any of them from dropping behind the others. Thus, the necessary requirement to this purpose is fast operation of all engaged components of the model as any such a delay disrupts the concurrency of the entire system, and adding NP computational time into the equation would itself render a solution suboptimal. Referring to Fig.10(a), it is clear that computational time for any component of the system and also whole operation of the system drawn in a reasonable interval in range of seconds and there is no remarkable difference between different combination of the proposed algorithms which confirms real-time performance of the proposed strategy and its capability in real time handling of task allocation/reallocation and dynamic changes of the operation terrain. This privilege is even further highlighted considering concurrency of the value of expected and path flight time in multiple re-planning process (in Fig.10(b)) in one mission for all proposed combination of algorithms. It is apparent in Fig.10(b) that expected time (presented by green boxplots for multiple recall in one mission) and path time (presented by blue compact boxplots for multiple recall in one mission) are lied in the close time range and there is no considerable difference between their variations that confirms concurrency of the components as the expected time is calculated from route time and path time should be accordant to the expected time, which is accurately satisfied in this research for all proposed algorithms that confirms robustness of the strategy regardless of type of applied meta-heuristic algorithm.

Fig.10(c) illustrates the comparison of elapsed time for carrying out a mission using different combination of algorithms. As mentioned earlier, it is highly critical to end the mission before vehicle runs out of battery and time. Also it is important for the model to tale maximum use of total time and finish the mission with minimum positive residual time. Considering the variations of residual time (yellow boxes), route(mission blue boxes) time, and total available time (red line over the boxes) in Fig.10(c), all missions terminated with a very small residual time and did not cross the predefined time boundary that confirms the remarkable capability of the proposed model in mission timing which is critical for fully autonomous operations; however, it seems the combination of ACO-DE, BBO-DE, BBO-BBO, and BBO-FA show the best timing performance between others.

Along with mentioned factors, the stability of the model in terms of managing total systems violation is also important issue that should be taken into consideration. According to results presented in Fig.11, the cost variation for the model and its components shows the stability of the model in producing optimal solutions as the cost variation range stands in a specific interval for all cases. The OPP is violated if the trajectory crosses the forbidden zones covered by coast or any obstacle; or when the vehicles surge, sway, yaw velocity egress the defined boundaries; or depth of the trajectory lies outside the predefined vertical borders; and the TAMP gets violation when the mission time exceeds the total mission available time. Both TAMP and OPP are recalled for several times in a one mission. Violation variations in Fig.11 shows that the model accurately satisfies all defined constraints and efficiently manages the engaged components toward eliminating the total violation regardless of the employed algorithms. The 150 Monte Carlo simulation runs are carried out for all applied algorithms to prove independency of the proposed strategy from applied metaheuristic methods, given by Fig.12 to Fig.15. Carrying out the Monte Carlo simulations gives more confidence about total model performance, stability and accuracy in dealing with random transformation of the environment or operational graph size and topology.

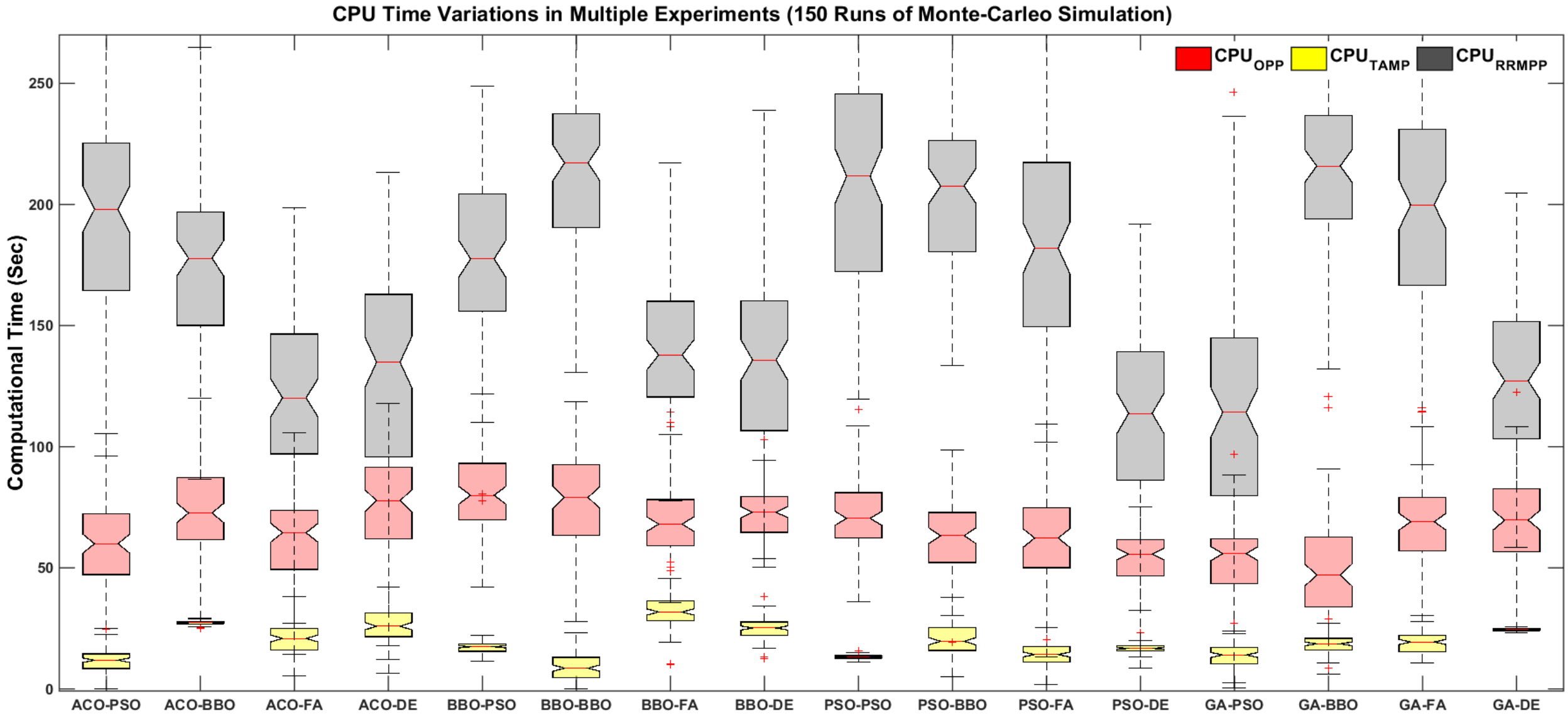

**Fig.12.** Statistical analysis computational performance of the $R^{II}MP^{II}$ model for multiple embedded meta-heuristic algorithms

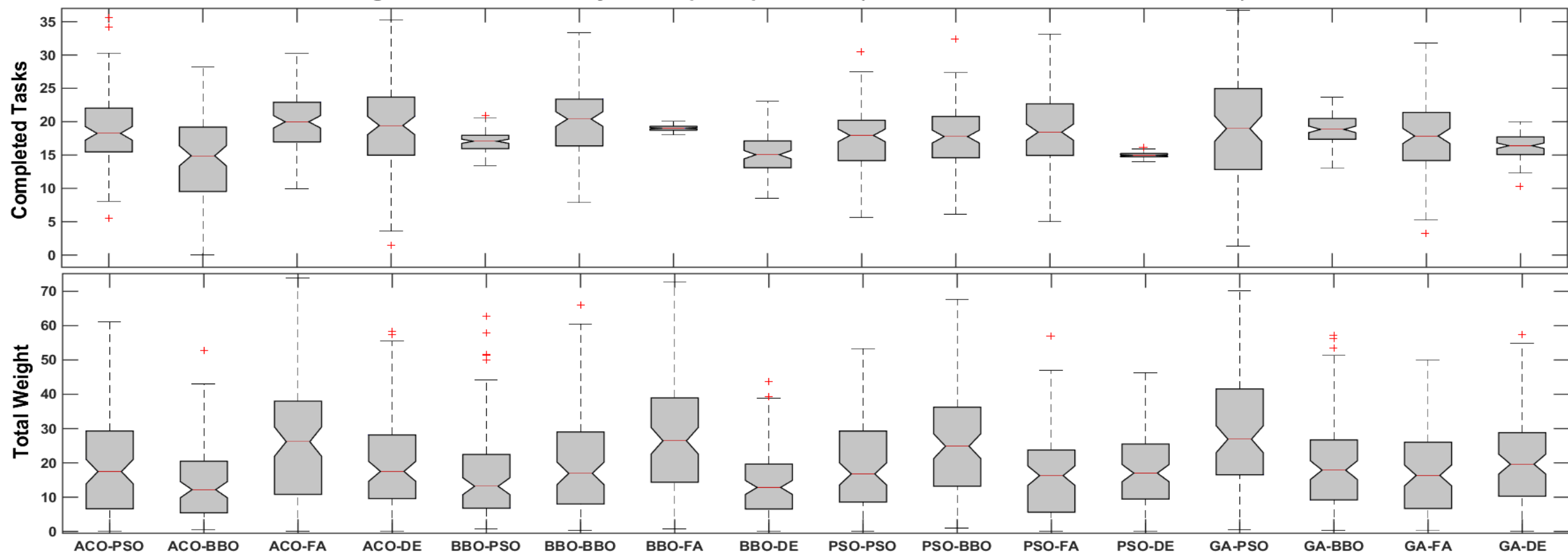

**Fig.13.** Average range of total obtained weight and number of completed tasks by different combination of embedded meta-heuristics in multiple experiments

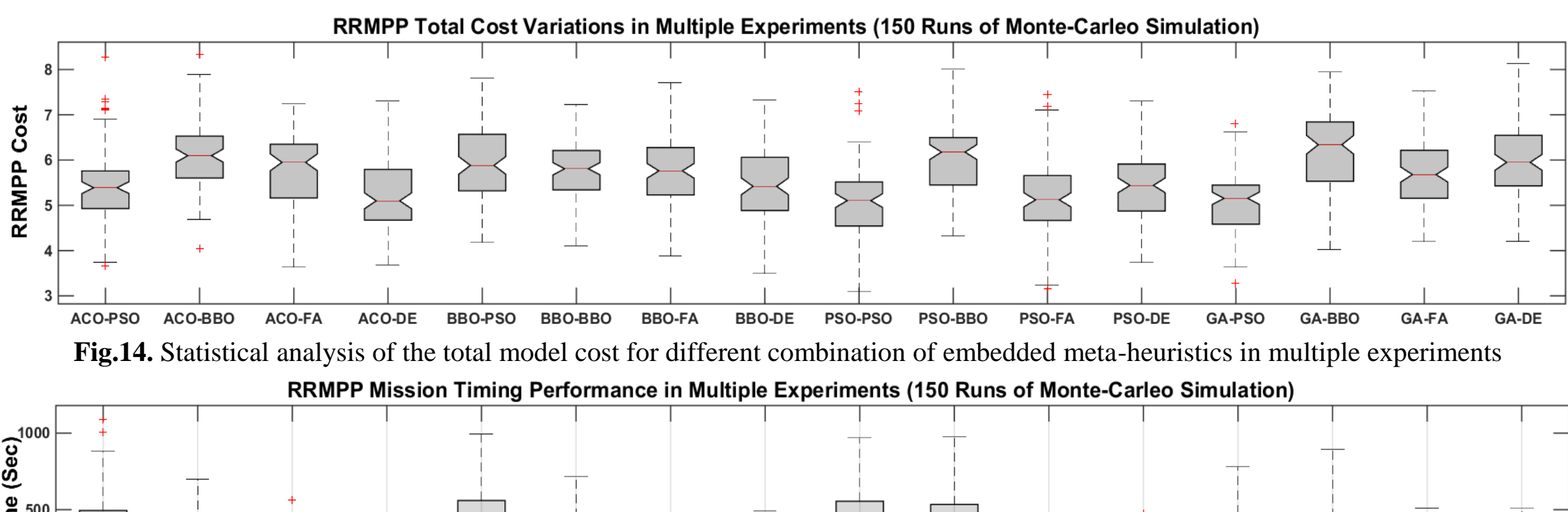

**Fig.14.** Statistical analysis of the total model cost for different combination of embedded meta-heuristics in multiple experiments

**Fig.15.** Average range of residual time variations in multiple experiments for different combination of embedded meta-heuristics

The results captured from Monte Carlo simulation in Fig.12 also confirms that the proposed model is remarkably appropriate for real-time application as the all CPU time variations are drawn in a very narrow boundary in range of seconds for all combination of employed algorithms by TAMP and OPP. It is also noteworthy to mention from the analysis of the results in Fig.13, the variation ranges of performance metrics of total obtained weight and completed tasks is almost in a same range for all applied algorithms that shows the stability and robustness of the model in maintaining beneficence of the mission in a reasonable range when dealing with environmental changes and random deformation of the topology of operation network. Referring to Fig.14, the total model cost is also varying in a similar range for all applied algorithms (almost between 4.5 to 7). The results obtained from analyze of cost variations in Fig.14 also demonstrate the inherent robustness and stability of the proposed model independent of the applied algorithm.

To ascertain the timing accuracy of the proposed $R^{II}MP^{II}$ model in different situations and against problem space deformations, variation range of the residual time in 150 Monte Carlo executions is presented by Fig.15 for all applied algorithms. It is noted from analyze of the results, the residual time variations for all cases is lied in a positive scale (except some outliers denoted by red cross sign under the boxplots) and comparing to the assigned time threshold of $T_{\Upsilon}$:$T_{Total}$=10800 (sec) considerably minimized and approached to zero which means maximum part of the available time is used. This achievement confirms the performance of the $R^{II}MP^{II}$ model in mission timing.

## 5 Conclusion

This paper looks at developing a scheme for high-level decision making of AUV missions by combining a reactive Task Assign Mission Planner (TAMP) and an Online Path Planner (OPP) that operate synchronously in a hierarchical architecture of $R^{II}MP^{II}$. Missions are assumed to be in the form of a distribution of tasks/waypoints, and the TAMP component choses the best sequence of tasks to be performed in a given time and the OPP guides the vehicle from one waypoint to the other. Various population based meta-heuristic approaches are tested in simulation and the stability and real-time applicability of the proposed $R^{II}MP^{II}$ framework is shown. The performance of TAMP is investigated applying ACO, BBO, PSO and GA algorithm and the OPP is validated using BBO, PSO, FA, and DE algorithms. The introduced OPP in the proceeding research is capable of using useful information of previous solutions to re-plan the subsequent trajectory in accordance with terrain changes and this considerably reduces the computational burden of the path planning process. Moreover, only local variations of the environment in vicinity of the vehicle is considered in path re-planning, so a small computational load is devoted for re-planning procedure since the upper layer of TAMP gives a general overview of the operation area that vehicle should fly thru. It is inferred from results of OPP simulation, the generated trajectory by all aforementioned algorithms efficiently respect environmental and vehicular constraints that means safety of vehicle's deployment is efficiently satisfied; while comparing the optimality of the solutions differs in different situations, but in conclusion FA acts more efficiently in satisfying vehicle path planning objectives in all examined cases.

After all, a systematic comparison of aforementioned meta-heuristics is carried out for evaluating the performance stability and robustness of the $R^{II}MP^{II}$ framework on mission timing, multitasking, safe and efficient deployment in a waypoint cluttered uncertain variable underwater environment. The simulation results carefully analyzed and discussed in Section 4. Accuracy and performance of all different combination of the employed algorithms has been studied globally in different situations that closely matches the real world conditions. This study aims to prove efficiency of meta-heuristics against any other deterministic methods for the purpose of this research rather than criticizing or comparing efficiency of the applied algorithms. Relying on analyze of the simulation results, all algorithms reveal very competitive performance in satisfying addressed performance metrics. Variation of the computational time for operation of all algorithms is settled in a narrow bound in range of second that confirm accuracy and real-time performance of the system. Independent of the type of meta-heuristic algorithm applied in this study, the proposed framework is able to achieve the mission objectives toward task-time management; and carrying out reliable, safe and beneficial deployment for a single vehicle operation in a restricted time.

This framework is a foundation for promoting vehicle's autonomy in different levels of self-awareness and mission management-scheduling, which are core elements of autonomous underwater missions.

**Acknowledgment**

Somaiyeh M.Zadeh and Amirmehdi Yazdani are funded by Flinders International Postgraduate Research Scholarship (FIPRS) program, Flinders University of South Australia. This research is also supported through a FIPRS scheme from Flinders University. The authors declare that there is no conflict of interest.